\documentclass[conference]{IEEEtran}
\usepackage{times}

\usepackage[numbers,sort&compress]{natbib}
\usepackage{multicol}
\usepackage[bookmarks=true]{hyperref}

\usepackage{algorithm}
\usepackage{algpseudocode}
\usepackage{xcolor}
\usepackage{amsmath,amssymb}

\usepackage{booktabs}  
\usepackage{multirow} 
\usepackage{array}    
\usepackage{setspace}
\usepackage{graphicx} 
\usepackage{tabularx}
\usepackage{colortbl}

\usepackage{balance}

\usepackage{capt-of}

\definecolor{bestcolor}{RGB}{230,245,230}  
\definecolor{legatocolor}{RGB}{240,248,255}  
\definecolor{headercolor}{RGB}{245,245,250}  
\definecolor{legatogray}{gray}{0.92}
\definecolor{legatoblue}{RGB}{225,239,255} 

\usepackage{cleveref}

\algrenewcommand\algorithmiccomment[1]{%
  \hfill{\color{black!55}\textit{$\triangleright$~#1}}%
}
\newcommand{\algblue}[1]{\textcolor{blue!70!green}{#1}}

\newcolumntype{Y}{>{\centering\arraybackslash}X}

\newcommand{\pmstd}[2]{#1 $\pm$ {\scriptsize #2}}

\usepackage{textcomp} 
\newcommand{\authdag}{\textsuperscript{\textdagger}}
\newcommand{\authddag}{\textsuperscript{\textdaggerdbl}}  
\IEEEoverridecommandlockouts

\begin{document}

\title{Learning Native Continuation for \\ Action Chunking Flow Policies}


\author{
\authorblockN{
Yufeng Liu\textsuperscript{1,2},
Hang Yu\textsuperscript{2,4},
Juntu Zhao\textsuperscript{1,2},
Bocheng Li\textsuperscript{2,5},
Di Zhang\textsuperscript{2,4},
Mingzhu Li\textsuperscript{2},
Wenxuan Wu\textsuperscript{2},\\
Yingdong Hu\textsuperscript{3},
Junyuan Xie\textsuperscript{2},
Junliang Guo\textsuperscript{2}\authddag,
Dequan Wang\textsuperscript{1}\authdag,
Yang Gao\textsuperscript{2,3}\authdag
}
\authorblockA{
\textsuperscript{1}Shanghai Jiao Tong University \quad
\textsuperscript{2}Spirit AI \quad
\textsuperscript{3}Tsinghua University \quad \\
\textsuperscript{4}Tongji University \quad
\textsuperscript{5}University of Science and Technology of China
}
Project page: \href{https://lyfeng001.github.io/Legato/}{lyfeng001.github.io/Legato/}
\thanks{This work was done during the internship at Spirit AI.\par
\authdag~Corresponding authors. \authddag~Project leader.}
}



%

\maketitle

\begin{abstract}
Action chunking enables Vision Language Action (VLA) models to run in real time, but naive chunked execution often exhibits discontinuities at chunk boundaries. Real-Time Chunking (RTC) alleviates this issue but is external to the policy, leading to spurious multimodal switching and trajectories that are not intrinsically smooth. We propose Legato, a training-time continuation method for action-chunked flow-based VLA policies. Specifically, Legato initializes denoising from a schedule-shaped mixture of known actions and noise, exposing the model to partial action information. Moreover, Legato reshapes the learned flow dynamics to ensure that the denoising process remains consistent between training and inference under per-step guidance. Legato further uses randomized schedule condition during training to support varying inference delays and achieve controllable smoothness. Empirically, Legato produces smoother trajectories and reduces spurious multimodal switching during execution, leading to less hesitation and shorter task completion time. Extensive real-world experiments show that Legato consistently outperforms RTC across five manipulation tasks, achieving approximately 10\% improvements in both trajectory smoothness and task completion time.
\end{abstract}



\section{Introduction}

Action chunking~\cite{lai2022action} has become a widely adopted strategy for deploying large Vision Language Action (VLA) models in real-world robotic systems~\cite{zitkovich2023rt,black2025pi_,kim2024openvla,wen2025tinyvla,zhao2023learning,Zhao2025CoTVLAVC,wen2025llada,yu2026twinbrainvla,yu2025point,zhao2025you}.
By predicting sequences of action vectors, chunking amortizes inference cost and enables high-frequency control.
However, naive chunked execution introduces a fundamental drawback: due to inference delay and the intrinsic multimodality of flow-based policies~\cite{lipman2022flow}, transitions between consecutive chunks are often not smooth, leading to visible discontinuities during execution.

\begin{figure}[t]
    \centering
    \includegraphics[width=1\linewidth]{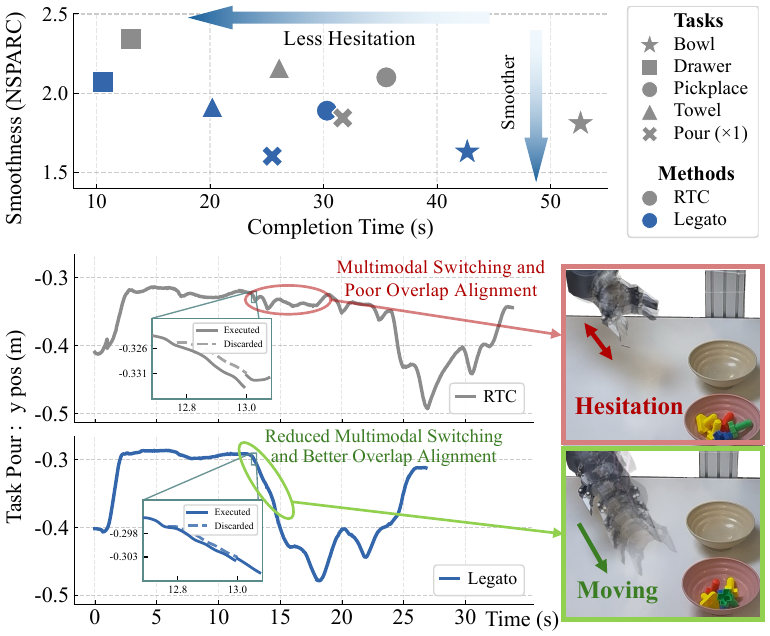}
    \caption{Legato reduces task completion time while improving trajectory smoothness compared to RTC~\cite{black2025real}.
Across five real-world manipulation tasks, Legato consistently achieves shorter execution time and lower NSPARC~\cite{Balasubramanian2015OnTA} (indicating smoother trajectories, discussed in \cref{sec:metrics}) than RTC.
The bottom plot shows an example execution trace on the \emph{pour} task, as defined in \cref{sec:tasks}, where Legato produces smoother action trajectories with fewer hesitation-induced slowdowns than RTC.}

    \label{fig:teaser}
\end{figure}

Real-Time Chunking (RTC)~\cite{black2025real} was proposed to mitigate this issue by applying inference-time inpainting~\cite{Pokle2023TrainingfreeLI,Song2023PseudoinverseGuidedDM} that partially constrains newly generated action chunks to previously generated actions in their overlapping regions.
While RTC improves continuity compared to naive chunked execution, its continuation mechanism is applied only at inference time and is not learned as part of the policy.
As a result, the policy is prone to spurious multimodal switching across chunk boundaries and producing trajectories that are not intrinsically smooth. 
Spurious multimodal switching often manifests as hesitation and abrupt direction changes, resulting in prolonged task completion time, as shown in \cref{fig:teaser}.

In this work, we argue that stable chunked execution requires chunk continuation to be a native, learned property of the policy.
Achieving this entails two requirements: \textbf{(i) per-step guidance}, where guidance is applied repeatedly across denoising steps, and \textbf{(ii) training-inference consistency}.
We propose Legato, a training-time continuation mechanism for action-chunked flow-based VLA policies.
Rather than learning the canonical flow-matching velocity field~\cite{lipman2022flow,black2024pi_0} and relying on inference-time correction, Legato internalizes chunk-to-chunk continuation into the learned denoising dynamics.

\begin{figure*}[t]
    \centering
    \includegraphics[width=1\linewidth]{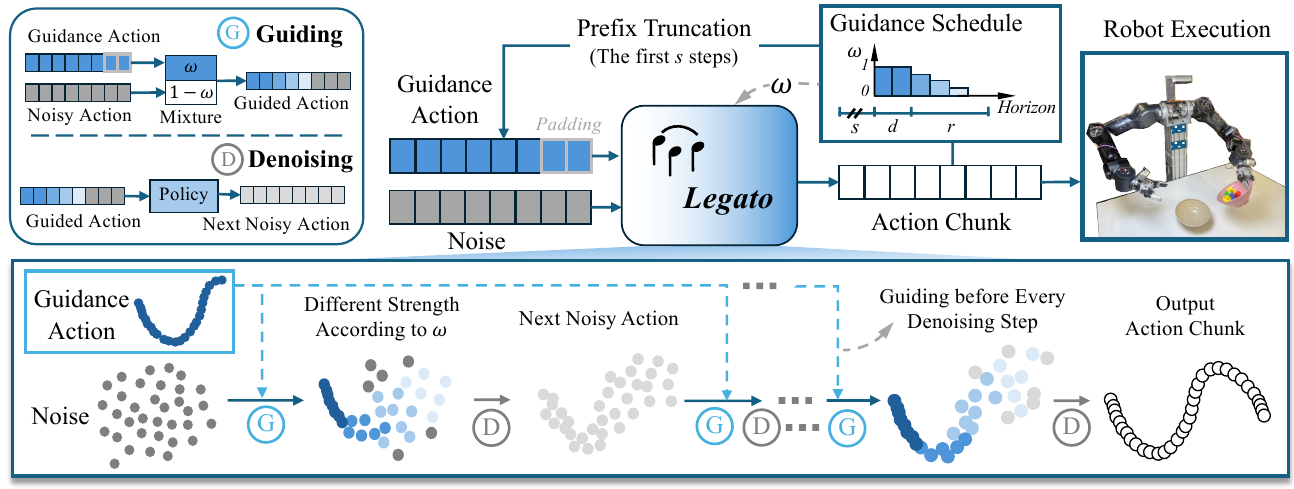}
    \caption{Overview of Legato with schedule-shaped continuation dynamics. The schedule parameters are defined as follows: $s$ is the executed length per cycle, $d$ sets the fully guided prefix (inference delay), and $r$ controls the ramp-down length of the guidance schedule over the remaining horizon. Given $\boldsymbol{\omega}$, Legato initializes actions via an action–noise mixture and learns a reshaped velocity field so that the native schedule effect is realized during multi-step denoising.}
    \label{fig:overview}
\end{figure*}

To satisfy requirement \emph{per-step guidance}, we first define a guidance schedule that specifies how strongly each timestep should adhere to the guidance actions.
Unlike Training-time RTC~\cite{black2025training}, which enforces continuation via a hard clamp on the prefix, Legato uses a smooth schedule: it anchors the beginning of the chunk to known actions and gradually ramps the guidance strength down to zero.
During training, the known actions are the ground-truth of the same chunk~\cite{black2025real}.
During inference, the known actions correspond to the overlapping prefix of the previously generated chunk.
This schedule-shaped design provides fine-grained control over the continuity strength between adjacent chunks.

With the schedule-shaped guidance, we enforce requirement \emph{training-inference consistency} under per-step guidance.
At inference time, action generation proceeds through multiple denoising steps, and empirically proved by \cref{subsec:motivation}, effective continuation requires per-step guidance before every denoising step.

Training-time RTC~\cite{black2025training} achieves this by hard-fixing the executed prefix and learning to denoise only the remaining horizon.
In contrast, Legato trains the policy to generate the entire chunk under per-step, schedule-shaped guidance by reshaping the velocity field.
This yields strict training-inference consistency, as shown in \cref{fig:overview}.

To make the above dynamics usable in real-world deployments, we account for variations in inference latency and desired continuation strength.
In real-world deployment, inference latency can vary across hardware and runtime optimizations~\cite{black2025real}.
Under a fixed guidance schedule, such variations lead to mismatched overlap regions and require retraining to maintain consistent behavior.
At the same time, we may want to adjust the schedule (i.e., ramp length) to control how strongly continuation is enforced.
To handle both factors, we randomize the schedule parameters during training and condition the policy on the resulting schedule, so the same model can adapt to different latencies and ramp lengths.

We evaluate Legato extensively in real-world environments to assess the necessity of learning action continuation as part of the policy dynamics.
We consider five diverse robotic manipulation tasks.
Across all settings, Legato consistently produces smoother trajectories and achieves significantly shorter task completion time by suppressing spurious multimodal switching compared to RTC, as shown in \cref{fig:teaser}.
Additional ablation studies further validate the robustness of Legato across different guidance schedules, VLA models, and conditioning strategies, demonstrating that its learned continuation behavior generalizes well under varying inference conditions.

Our work offers three main contributions:
\begin{itemize}
    \item We propose Legato, a training-time continuation framework that enables per-step, schedule-shaped guidance while maintaining strict training-inference consistency by reshaping the flow dynamics of action-chunked policies.
    \item We introduce randomized schedule conditioning to support varying inference delays and to provide flexible control over trajectory smoothness.
    \item Extensive real-robot experiments across five manipulation tasks show that Legato consistently outperforms RTC and training-time RTC, producing smoother trajectories and shorter task completion time.
\end{itemize}
\section{Related Works}

\subsection{VLA and Action Chunking Methods}
Recent Vision Language Action (VLA) models couple large vision–language representations with learned heads to enable end-to-end visuomotor policies~\cite{black2024pi_0,black2025pi_,bjorck2025gr00t,liu2024rdt,Lin2025OneTwoVLAAU,team2024octo,team2025gemini,cheang2025gr,Wang2025VQVLAIV,zheng2024tracevla,zhen20243d}. Most VLA systems generate actions in chunks, predicting a sequence of future controls per inference step~\cite{Zhao2025CoTVLAVC,Wang2025VQVLAIV,qu2025eo}. Action chunking has been successfully combined with a variety of generative policy formulations, including diffusion-based~\cite{chi2025diffusion,liang2025discrete,wen2025dvla,wen2025llada,pearce2023imitating,chi2024universal,barreiros2025careful}, flow-based~\cite{black2024pi_0,black2025pi_,lin2025evo,braun2024riemannian}, and discrete action representations~\cite{pertsch2025fast,belkhale2024minivla}. However, chunked execution trades off responsiveness for smoothness~\cite{black2025real}, and inference latency further increases discontinuities between successive chunks, motivating methods to improve continuation.

\subsection{Trajectory Continuation in Learned Policies}
Building on action chunking, a common approach to improve responsiveness is asynchronous execution, where action generation overlaps with execution~\cite{zhao2023learning,shukor2025smolvla}; however, without explicit continuation constraints, independently generated chunks can exhibit abrupt multimodal switches at their boundaries. Bidirectional decoding (BID)~\cite{liu2024bidirectional} uses rejection sampling to keep continuity across chunks. Real-time chunking (RTC)~\cite{black2025real} addresses continuation under asynchronous inference by conditioning new action chunks on previously issued actions that are guaranteed to execute. While RTC effectively mitigates boundary artifacts caused by inference latency, it is an inference-time mechanisms, leaving open the question of how to induce robust trajectory continuation without additional test-time intervention.


\subsection{Conditioning in Diffusion- and Flow-Based Policies}
Recent diffusion-~\cite{ho2020denoising} and flow-based~\cite{lipman2022flow} policies explore conditioning mechanisms to improve temporal coherence and execution efficiency.
Diffusion Forcing~\cite{chen2024diffusion} and Fast Policy Synthesis with Variable Noise Diffusion Models~\cite{hoeg2024streaming} adopt a timestep-level diffusion formulation, generating a single action per inference step and improving reactivity through noise modulation, but without explicitly modeling continuation across action chunks.
Rolling Diffusion Policy~\cite{jung2025rolling} similarly operates at the timestep level, incrementally refining future actions via rolling denoising to enhance temporal awareness.
In contrast, SAIL~\cite{arachchige2025sail} performs chunk-level conditioning by leveraging overlapping actions between consecutive chunks using classifier-free guidance~\cite{ho2022classifier}, which mitigates discontinuities under fast execution but provides only soft alignment and limited control over continuation strength.
Concurrent with our work, training-time RTC~\cite{black2025training} introduces continuation during training by conditioning on a hard action prefix that simulates inference delay.
While this exposes the policy to prefix-based continuation, the conditioning remains an external constraint and does not account for the effective denoising dynamics induced by repeated, schedule-shaped guidance at inference time, leaving continuation outside the learned policy dynamics.

\section{Methodology}

\label{sec:method}

\subsection{Preliminaries}
\label{sec:preliminaries}

We consider Vision Language Action (VLA) policies that generate action sequences in fixed-length chunks using flow-based generative models.
Let $\mathbf{A}\in\mathbb{R}^{H\times D_a}$ denote a ground-truth action chunk of horizon $H$, where $D_a$ is the action dimension, and let $\boldsymbol{\epsilon} \sim \mathcal{N}(\mathbf{0}, \mathbf{I})$ denote Gaussian noise of the same shape.
Flow matching (FM)~\cite{lipman2022flow} constructs a continuous-time interpolation between noise and action, and trains a neural velocity field to transport samples along this path.

\subsubsection{\textbf{Flow matching}}
Given a time variable $t \in [0,1]$, standard flow matching defines the interpolation
\begin{equation}
\label{eq:fm_path}
\mathbf{X}_t = (1-t)\,\boldsymbol{\epsilon} + t\,\mathbf{A},
\end{equation}
and supervises the model to predict the corresponding velocity field
\begin{equation}
\label{eq:fm_velocity}
\mathbf{u}^{\mathrm{FM}}(\mathbf{X}_t, t) = \mathbf{A} - \boldsymbol{\epsilon}.
\end{equation}
At inference time, action generation begins from an initial noise sample $\boldsymbol{\epsilon}$ and progressively transforms it into an action chunk by integrating the learned velocity field from $t=0$ to $t=1$.

\begin{figure}[t]

\centering
\includegraphics[width=1\linewidth]{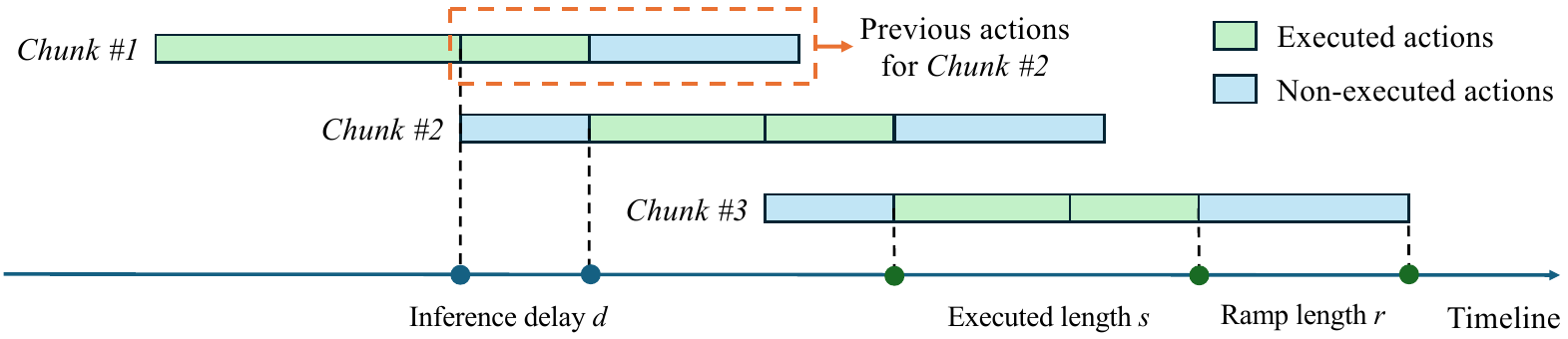}
\caption{Continuous chunk execution and guidance construction: $d$ (full guidance), $r$ (ramp-down), $s$ (executed length), $r{+}s{+}d{=}H$.}
\vspace{-3pt}
\label{fig:timeline}
\end{figure}

\subsubsection{\textbf{Real-Time Chunking}}
Real-Time Chunking (RTC)~\cite{black2025real} enforces continuity between successive action chunks through a test-time guidance mechanism inspired by inpainting, which encourages partial agreement with previously generated actions.

Beyond continuity, RTC also introduces an asynchronous execution scheme that overlaps inference and action execution to mitigate model latency.
For an action chunk of horizon $H$, the first $d$ timesteps correspond to inference latency, during which the robot continues executing the previous chunk.
The next $s$ timesteps correspond to the portion of the current chunk that will be executed before the next inference completes.
Once $(s-d)$ timesteps of this portion have been executed, inference for the next chunk is triggered while execution continues, enabling overlapped computation and control.

RTC further employs a structured guidance schedule over the chunk horizon.
The initial $d$ timesteps receive full guidance to strictly enforce continuity with past actions, followed by a ramp-down phase.
Let $r$ denote the length of this ramp; the schedule commonly satisfies
\begin{equation}
\label{eq:rtc_schedule}
r + s + d = H .
\end{equation}
This design enforces strong adherence to previously executed actions near the chunk boundary while gradually relaxing constraints toward the end of the horizon.

Our method draws inspiration from RTC in both its use of previously executed actions and its structured guidance scheduling, as shown in \cref{fig:timeline}, green regions indicate actual execution.

\subsection{Why per-step guidance matters?}
\label{subsec:motivation}

We aim to learn a policy that remains strictly consistent between training and inference.
To satisfy this requirement, guidance must be incorporated during training.
Under the standard FM formulation, training optimizes the velocity field that transports samples from an initial noise to the ground-truth action.
The only inference strategy that remains strictly consistent with training is to apply guidance only once at initialization and then perform multi-step denoising.

To verify whether the one-shot guidance is enough for continuous guidance, we trained a flow policy where the standard noise initialization was replaced with a prefix-guided variant $\boldsymbol{\epsilon}'$.
Let $\mathbf{m}\in\{0,1\}^H$ denote a horizon-wise mask for the overlap region, and let $\mathbf{A}_{\mathrm{ref}}$ be the ground-truth reference actions on this overlap.
We construct the guided noise:
\begin{equation}
\label{eq:oneshot_eps_prime}
\boldsymbol{\epsilon}'
= (\mathbf{1}-\mathbf{m})\odot \boldsymbol{\epsilon}
+ \mathbf{m}\odot \mathbf{A}_{\mathrm{ref}},
\
\boldsymbol{\epsilon}\sim\mathcal{N}(\mathbf{0},\mathbf{I}),
\end{equation}
where $\odot$ denotes element-wise multiplication.
Training follows standard flow matching, but with $\boldsymbol{\epsilon}'$ as the start point:
\begin{equation}
\label{eq:oneshot_path}
\mathbf{X}_t = (1-t)\,\boldsymbol{\epsilon}' + t\,\mathbf{A},
\
\mathbf{u}^{\mathrm{FM}}(\mathbf{X}_t,t)=\mathbf{A}-\boldsymbol{\epsilon}'.
\end{equation}
During inference, we apply the same prefix clamp only at initialization, $\mathbf{X}_0=\boldsymbol{\epsilon}'$, and then perform standard multi-step denoising without any intermediate guidance.

However, empirical results reveal that one-shot guidance is insufficient for continuous guidance.
As the denoising process iterates, the overlap region in the generated chunk progressively deviates from the reference actions as shown in \cref{fig:denoise} (details are shown in the appendix).
The prefix part moves away from the desired constraint without repeated guidance. 

This study leads to a crucial conclusion: \textbf{effective continuation requires per-step guidance.}
However, simply applying per-step guidance on the standard FM remains inconsistent with the training objective.
This dilemma motivates Legato:
we reshape the flow dynamics during training so that the model can support per-step, schedule-shaped guidance while remaining fully consistent with the training objective.

\begin{figure}[t]
    \centering
    \includegraphics[width=1\linewidth]{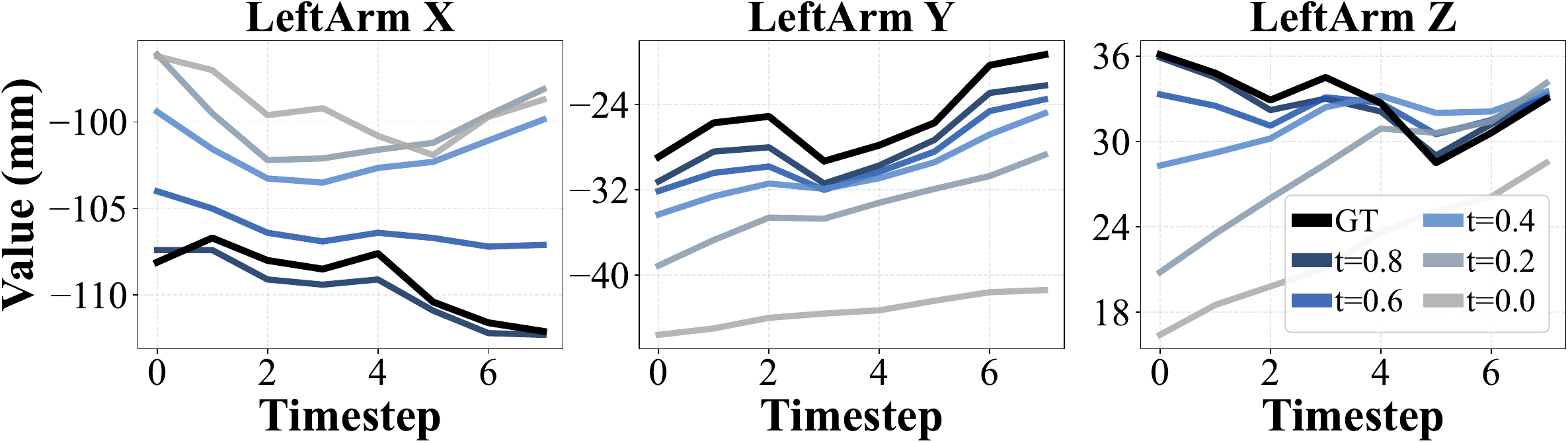}

\caption{
One-shot prefix guidance cannot preserve prefix constraints during denoising.
Trajectories show three dimensions of the overlap (prefix) actions across denoising steps; colors indicate diffusion times $t$ (from $1$ to $0$), and GT denotes the ground-truth prefix.
Although clamped at initialization, the overlap actions drift from the reference as denoising proceeds, motivating the need for per-step guidance.
Evaluated on the \textit{pour} task, as defined in \cref{sec:tasks}.
}
    \label{fig:denoise}
    
\end{figure}

\subsection{Native Continuation for Action Chunk Generation}
\label{sec:native_continuation}
We aim to make action continuation a \emph{native property} of the learned policy.
This entails two requirements: \textbf{(i) per-step guidance} as discussed in \cref{subsec:motivation}, where guidance is applied repeatedly across denoising steps, and \textbf{(ii) training-inference consistency}.

Accordingly, we first construct a schedule-shaped training path, then derive the induced guided dynamics, and finally reshape the velocity field to eliminate the train-test mismatch.

\subsubsection{\textbf{Action-noise mixture}}
To incorporate guidance into training, we introduce a horizon-wise continuation vector $\boldsymbol{\omega} \in [0,1]^H$, which encodes the guidance schedule over the chunk horizon, i.e., full guidance near the chunk beginning and a gradual ramp-down toward the end of the horizon.

Using $\boldsymbol{\omega}$, we define an action-noise mixture
\begin{equation}
\label{eq:noise_eff}
\boldsymbol{\epsilon}_{\mathrm{eff}}
= (\mathbf{1}-\boldsymbol{\omega}) \odot \boldsymbol{\epsilon}
+ \boldsymbol{\omega} \odot \mathbf{A},
\end{equation}
where $\odot$ denotes element-wise multiplication and
$\boldsymbol{\epsilon}_{\mathrm{eff}}$ represents the \emph{effective noise initialization} induced by continuation guidance, interpolating between the action chunk and pure noise in a horizon-wise manner.

Based on this mixture, we construct the interpolation path
\begin{equation}
\label{eq:legato_path}
\mathbf{Y}_t
= (1-t)\,\boldsymbol{\epsilon}_{\mathrm{eff}} + t\,\mathbf{A},
\end{equation}
which reduces to the standard flow-matching path when $\boldsymbol{\omega}=\mathbf{0}$ and collapses to the action chunk for all $t$ when $\boldsymbol{\omega}=\mathbf{1}$.

The corresponding flow-matching velocity is
\begin{equation}
\label{eq:legato_teacher_fm}
\mathbf{u}^{\mathrm{FM}}(\mathbf{Y}_t,t)
= \mathbf{A} - \boldsymbol{\epsilon}_{\mathrm{eff}}
= (\mathbf{1}-\boldsymbol{\omega}) \odot (\mathbf{A}-\boldsymbol{\epsilon}),
\end{equation}
reflecting a horizon-wise modulation of the FM velocity.

\begin{algorithm}[t]
\caption{Legato: Training and Inference}
\label{alg:legato}
\begin{algorithmic}[1]
\Require
policy $f_\theta$; observation $o$; horizon $H$; denoising steps $N$; schedule params $(d,r)$; executed length $s$
\State Construct schedule $\boldsymbol{\omega}\in[0,1]^H$ from $(d,r)$
\State $\Delta t \gets 1/N$, \quad $\boldsymbol{\kappa} \gets \boldsymbol{\omega}/\Delta t$

\If{training}
    \State Sample $t\sim\mathcal{U}(0,1)$,
    $\boldsymbol{\epsilon}\sim\mathcal{N}(\mathbf{0},\mathbf{I})$
    \State $\boldsymbol{\epsilon}_{\mathrm{eff}}
    \gets \boldsymbol{\omega}\odot\mathbf{A}
    + (\mathbf{1}-\boldsymbol{\omega})\odot\boldsymbol{\epsilon}$
    \State $\mathbf{Y}_t \gets (1-t)\boldsymbol{\epsilon}_{\mathrm{eff}} + t\mathbf{A}$

    \State \algblue{$\mathbf{v}_{\text{target}}
    \gets (1-\boldsymbol{\kappa}\odot(1-t))\odot(\mathbf{A}-\boldsymbol{\epsilon})$}

    \State \algblue{Update $\theta$ by
    $\|f_\theta(\mathbf{Y}_t,o,t,\boldsymbol{\omega})-\mathbf{v}_{\text{target}}\|_2^2$}

\Else \Comment{Inference}
    \State Sample $\boldsymbol{\epsilon}\sim\mathcal{N}(\mathbf{0},\mathbf{I})$
    \If{no previous chunk}
        \State $\mathbf{A}_{\mathrm{prev}} \gets \mathbf{0}$
    \EndIf
    \State $\mathbf{A}_{\mathrm{ref}}
    \gets \textsc{PadLast}\!\left(\mathbf{A}_{\mathrm{prev}}[s{:}H]\right)$
    
    \Comment{Truncate and pad with the last value to length $H$}

    \State $\mathbf{X}_0 \gets \boldsymbol{\epsilon}$

    \For{$k=0$ \textbf{to} $N-1$}
        \State \algblue{$\mathbf{Y}_k \gets (\mathbf{1}-\boldsymbol{\omega})\odot \mathbf{X}_k
        + \boldsymbol{\omega}\odot \mathbf{A}_{\mathrm{ref}}$}
        \Comment{Guiding}
        \State \algblue{$\mathbf{X}_{k+1} \gets \mathbf{Y}_k
        + \Delta t\, f_\theta(\mathbf{Y}_k,o,t_k,\boldsymbol{\omega})$}
        \Comment{Denoising}
    \EndFor
    \State \Return $\hat{\mathbf{A}}\gets \mathbf{X}_N$
\EndIf
\end{algorithmic}
\end{algorithm}

\textbf{Multimodal persistence and smoothness:}
Eq.~\eqref{eq:legato_teacher_fm} reveals a \emph{schedule-shaped} velocity: the target transport magnitude is modulated by $\boldsymbol{\omega}$.
In timesteps with large $\boldsymbol{\omega}_i$ (strong continuation), the effective $u^{\mathrm{FM}}_i$ is suppressed as $u^{\mathrm{FM}}_i \propto (1-\boldsymbol{\omega}_i)$, making the overlap  and the ramp region intrinsically less mutable than the none-guidance region during denoising.
This discourages frequent switching among competing action modes in highly multimodal tasks.

Moreover, since $\boldsymbol{\omega}$ decreases from 1 along the horizon, the effect of continuation is gradually relaxed through a ramp, yielding a smooth transition from strict guidance to free generation.
As a result, Legato reduces chunk-boundary discontinuities and improves trajectory smoothness.

\subsubsection{\textbf{Effective dynamics of repeated continuation guidance}}

The velocity construction above specifies the schedule-shaped guidance.
At inference time, continuation requires per-step guidance to keep effective.
We therefore derive the exact dynamics induced by the per-step guidance.

At each denoising step $k$, the current noisy action is first guided toward the reference action according to the guidance schedule $\boldsymbol{\omega}$:
\begin{equation}
\label{eq:guided_state}
\mathbf{Y}_k
= (\mathbf{1}-\boldsymbol{\omega}) \odot \mathbf{X}_k
+ \boldsymbol{\omega} \odot \mathbf{A},
\end{equation}
where $\mathbf{A}$ denotes the reference action and $\mathbf{X}_k$ is the current noisy action before guidance.

We then perform one denoising update:
\begin{equation}
\label{eq:guided_update}
\mathbf{X}_{k+1}
= \mathbf{Y}_k + \Delta t\, f_\theta(\mathbf{Y}_k,t_k),
\end{equation}
after which the same guidance in \cref{eq:guided_state} is applied again at the next step, as shown in \cref{fig:overview}.

Eliminating $\mathbf{X}_k$ yields the exact recurrence
\begin{equation}
\label{eq:y_recurrence}
\mathbf{Y}_{k+1}
= \boldsymbol{\omega} \odot \mathbf{A}
+ (\mathbf{1}-\boldsymbol{\omega}) \odot \mathbf{Y}_k
+ (\mathbf{1}-\boldsymbol{\omega}) \odot \Delta t\, f_\theta(\mathbf{Y}_k,t_k).
\end{equation}
Taking the continuous-time limit, this recurrence corresponds to the ordinary differential equation
\begin{equation}
\label{eq:guided_ode}
\dot{\mathbf{Y}}(t)
= (\mathbf{1}-\boldsymbol{\omega}) \odot f_\theta(\mathbf{Y}(t),t)
- \boldsymbol{\kappa} \odot (\mathbf{Y}(t)-\mathbf{A}),
\
 \boldsymbol{\kappa} = \boldsymbol{\omega}/\Delta t .
\end{equation}
Importantly, \cref{eq:guided_ode} is not an approximation: it is the exact continuous-time system whose Euler discretization reproduces repeated continuation guidance.

\subsubsection{\textbf{Training-inference consistency}}
Having characterized the dynamics induced by per-step guidance, we now turn to the second requirement: training-inference consistency.
Standard flow matching supervises the velocity field $\mathbf{u}^{\mathrm{FM}}$, whereas inference with repeated continuation guidance follows the dynamics in \cref{eq:guided_ode}.
To eliminate this mismatch, we require the executed velocity field to coincide with the flow-matching target:
\begin{equation}
\label{eq:consistency_condition}
(\mathbf{1}-\boldsymbol{\omega}) \odot f_\theta(\mathbf{Y},t)
- \boldsymbol{\kappa} \odot (\mathbf{Y}-\mathbf{A})
= \mathbf{u}^{\mathrm{FM}}(\mathbf{Y},t).
\end{equation}
Solving \cref{eq:consistency_condition} for $f_\theta$ yields the \emph{Legato velocity field}
\begin{equation}
\label{eq:legato_velocity}
f_\theta(\mathbf{Y},t)
= (\mathbf{1}-\boldsymbol{\omega})^{-1} \odot
\bigl[
\mathbf{u}^{\mathrm{FM}}(\mathbf{Y},t)
+ \boldsymbol{\kappa} \odot (\mathbf{Y}-\mathbf{A})
\bigr],
\end{equation}
where the inverse is taken element-wise.

Substituting \cref{eq:legato_path} and \cref{eq:legato_teacher_fm} into \cref{eq:legato_velocity}, we obtain a closed-form target velocity
\begin{equation}
\label{eq:legato_target}
\mathbf{v}_{\text{target}}(t,\mathbf{A},\boldsymbol{\epsilon},\boldsymbol{\omega})
=
\bigl(1 - \boldsymbol{\kappa} \odot (1-t)\bigr)
\odot (\mathbf{A}-\boldsymbol{\epsilon}),
\end{equation}
The network is trained by regressing $f_\theta(\mathbf{Y}_t,o,t,\boldsymbol{\omega})$ to $\mathbf{v}_{\text{target}}$.
Thus, Legato preserves the geometric direction of standard flow matching while reshaping the velocity magnitude to internalize continuation dynamics.

{\textbf{Inference:}}
At inference time, we use the previously generated (but haven't been executed) chunk as the reference for continuation.
We construct a reference action chunk $\mathbf{A}_{\mathrm{ref}}$ from the previous prediction using the alignment procedure as shown in \cref{alg:legato}.
We then instantiate the guidance term in \cref{eq:guided_ode} by setting $\mathbf{A}\leftarrow \mathbf{A}_{\mathrm{ref}}$.

Given a schedule $\boldsymbol{\omega}$, we initialize
\begin{equation}
\label{eq:infer_init}
\mathbf{Y}_0
= \boldsymbol{\omega}\odot \mathbf{A}_{\mathrm{ref}}
+ (\mathbf{1}-\boldsymbol{\omega}) \odot \boldsymbol{\epsilon},
\
\boldsymbol{\epsilon}\sim\mathcal{N}(\mathbf{0},\mathbf{I}),
\end{equation}
We integrate \cref{eq:guided_ode} forward in time from $t=0$ to $t=1$ using the learned velocity field in \cref{eq:legato_velocity}, with the same discretization (number of denoising steps $N$) as used during training. This enable the strict training-inference alignment.

\subsection{Schedule Randomization and Conditioning}
\label{sec:parameter_randomization}

In our framework, the continuation schedule over an action chunk of horizon $H$ is fully specified by two scalar parameters: the inference delay $d$ and the ramp length $r$.
Given $(d,r)$, the guidance schedule $\boldsymbol{\omega}\in[0,1]^H$ is uniquely determined, consisting of a full-guidance prefix of length $d$ followed by a ramp part of length $r$.

In real-world deployment, effective inference delay varies across hardware platforms, model sizes, and inference optimizations.
To account for this variability while enabling flexible control over continuation smoothness, we randomize $(d,r)$ during training, thereby exposing the policy to a diverse family of guidance schedules.

When training with randomized schedules, the policy must be informed of the schedule at inference time.
We therefore explicitly condition the action decoder on the schedule.
Concretely, if the noisy action $\mathbf{Y}_t \in \mathbb{R}^{H\times D_a}$, where $D_a$ denotes the action dimension, we append the guidance schedule along the feature dimension, resulting in an noisy action of shape $(H, D_a + 1)$.
At inference time, adapting to a new continuation regime only requires changing the guidance schedule $\boldsymbol{\omega}$, without retraining the model.
Empirically, this schedule conditioning substantially improves robustness across hardware platforms and inference budgets.

\section{Experiments}

\subsection{Experimental Setups}

\begin{figure}[t]
    \centering
    \includegraphics[width=1\linewidth]{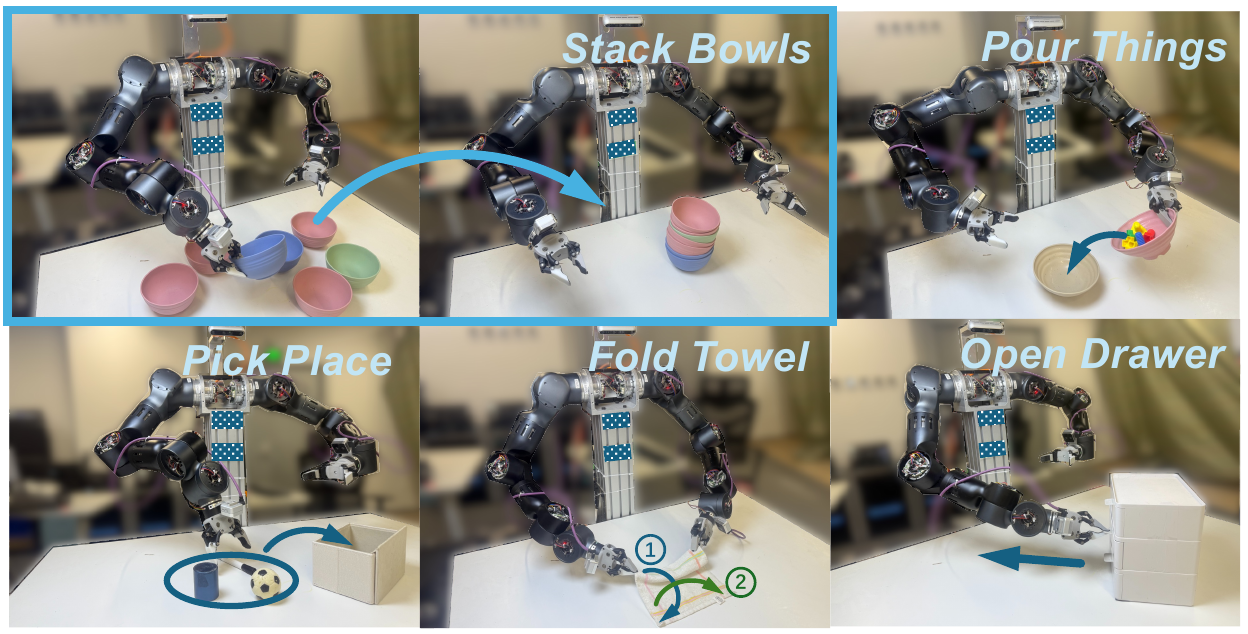}
    \caption{Real-world evaluation tasks on a dual-arm robot.
We consider five manipulation tasks (stack bowls, pour things, pick and place, fold towel and open drawer) covering diverse motion patterns and multimodal choices such as alternative grasp goals and left/right arm selection.}
    \label{fig:env}
\end{figure}


\begin{table*}[t]
\centering
\caption{Main real-world results comparing RTC and Legato across five tasks.
We report task score ($\uparrow$), completion time in seconds ($\downarrow$), and smoothness metrics ($\downarrow$): NLDLJ (Negative Log Dimensionless Jerk~\cite{arachchige2025sail,Balasubramanian2015OnTA}), NSPARC (Negative Linear and Angular Spectral Arc Length~\cite{arachchige2025sail,Balasubramanian2015OnTA}), and overlap RMSE (Root Mean Squared Error, $\times 10^3$).
Values are reported as mean $\pm$ standard error.}
\label{tab:real_world_main_results}

\renewcommand{\arraystretch}{0.9}
\setlength{\tabcolsep}{5pt}

\large
\resizebox{\textwidth}{!}{%
\begin{tabular}{l|cc|cc|cc|cc|cc}
\toprule
\multirow{3}{*}[-0.8em]{\textbf{Task}} 
& \multicolumn{2}{c|}{\multirow{2}{*}[-2pt]{\textbf{Score} $\uparrow$}} 
& \multicolumn{2}{c|}{\multirow{2}{*}[-2pt]{\textbf{Completion Time (s)} $\downarrow$}} 
& \multicolumn{6}{c}{\textbf{Smoothness} $\downarrow$} \\ 

\cmidrule(lr){6-11}

& \multicolumn{2}{c|}{} 
& \multicolumn{2}{c|}{} 
& \multicolumn{2}{c}{\textbf{NLDLJ} $\downarrow$}
& \multicolumn{2}{c}{\textbf{NSPARC} $\downarrow$}
& \multicolumn{2}{c}{\textbf{Overlap RMSE} ($\times 10^{3}$) $\downarrow$} \\

\cmidrule(lr){2-3}\cmidrule(lr){4-5}\cmidrule(lr){6-7}\cmidrule(lr){8-9}\cmidrule(lr){10-11}

& RTC & \cellcolor{legatoblue}{Legato}
& RTC & \cellcolor{legatoblue}{Legato}
& RTC & \cellcolor{legatoblue}{Legato}
& RTC & \cellcolor{legatoblue}{Legato}
& RTC & \cellcolor{legatoblue}{Legato} \\
\midrule

Bowls
& \pmstd{8.68}{0.35} & \cellcolor{legatoblue}\textbf{\pmstd{9.08}{0.33}}
& \pmstd{52.88}{3.54} & \cellcolor{legatoblue}\textbf{\pmstd{42.66}{2.68}}
& \pmstd{36.00}{0.34} & \cellcolor{legatoblue}\textbf{\pmstd{35.86}{0.38}}
& \pmstd{1.82}{0.04} & \cellcolor{legatoblue}\textbf{\pmstd{1.63}{0.02}}
& \pmstd{6.83}{0.50} & \cellcolor{legatoblue}\textbf{\pmstd{4.58}{0.17}} \\

Pour
& \pmstd{9.34}{0.18} & \cellcolor{legatoblue}\textbf{\pmstd{9.72}{0.13}}
& \pmstd{95.07}{2.86} & \cellcolor{legatoblue}\textbf{\pmstd{75.73}{1.51}}
& \pmstd{39.82}{0.15} & \cellcolor{legatoblue}\textbf{\pmstd{39.50}{0.13}}
& \pmstd{2.85}{0.24} & \cellcolor{legatoblue}\textbf{\pmstd{1.65}{0.08}}
& \pmstd{7.64}{0.70} & \cellcolor{legatoblue}\textbf{\pmstd{5.14}{0.17}} \\

PickPlace
& \pmstd{9.47}{0.15} & \cellcolor{legatoblue}\textbf{\pmstd{9.53}{0.12}}
& \pmstd{35.53}{1.24} & \cellcolor{legatoblue}\textbf{\pmstd{30.37}{0.65}}
& \pmstd{34.42}{0.18} & \cellcolor{legatoblue}\textbf{\pmstd{34.34}{0.14}}
& \pmstd{2.10}{0.08} & \cellcolor{legatoblue}\textbf{\pmstd{1.89}{0.05}}
& \pmstd{10.17}{0.66} & \cellcolor{legatoblue}\textbf{\pmstd{5.98}{0.40}} \\

Drawer
& \pmstd{9.20}{0.16} & \cellcolor{legatoblue}\textbf{\pmstd{9.50}{0.13}}
& \pmstd{25.97}{0.74} & \cellcolor{legatoblue}\textbf{\pmstd{21.80}{0.72}}
& \pmstd{32.73}{0.13} & \cellcolor{legatoblue}\textbf{\pmstd{28.55}{0.26}}
& \pmstd{2.24}{0.05} & \cellcolor{legatoblue}\textbf{\pmstd{1.99}{0.08}}
& \pmstd{12.11}{0.66} & \cellcolor{legatoblue}\textbf{\pmstd{11.74}{0.55}} \\

Towel
& \pmstd{7.33}{0.62} & \cellcolor{legatoblue}\textbf{\pmstd{8.17}{0.56}}
& \pmstd{25.93}{0.98} & \cellcolor{legatoblue}\textbf{\pmstd{20.00}{0.78}}
& \pmstd{32.79}{0.20} & \cellcolor{legatoblue}\textbf{\pmstd{32.43}{0.24}}
& \pmstd{2.17}{0.07} & \cellcolor{legatoblue}\textbf{\pmstd{1.97}{0.05}}
& \pmstd{11.28}{0.55} & \cellcolor{legatoblue}\textbf{\pmstd{6.22}{0.66}} \\

\bottomrule
\end{tabular}
}
\end{table*}
\begin{figure*}[t]
    \centering
    \includegraphics[width=1\linewidth]{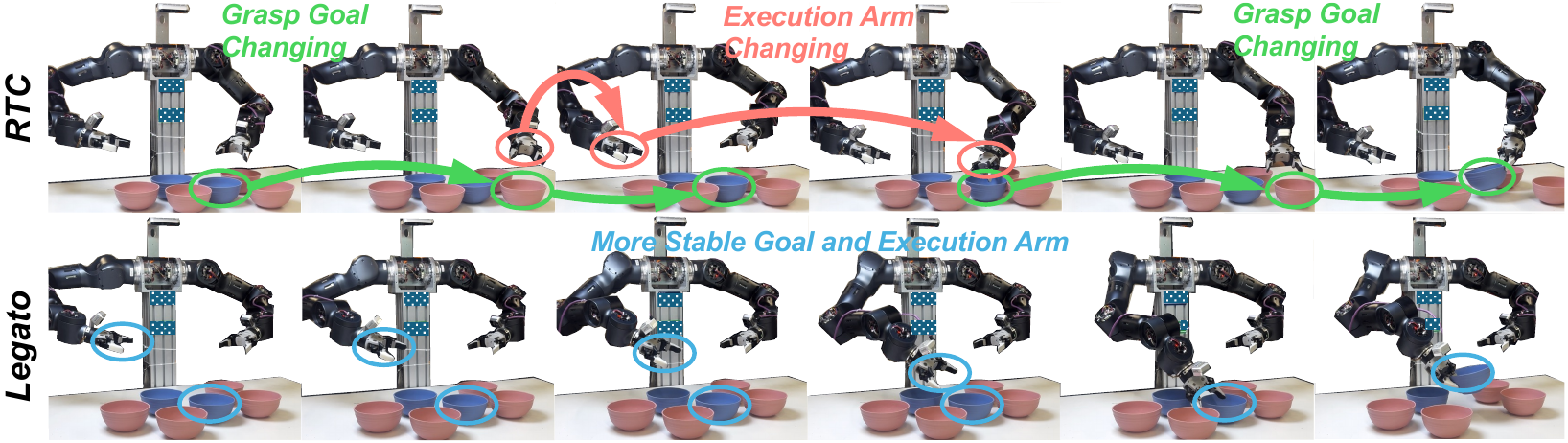}
    \vspace{-2mm} 
    \caption{Legato suppresses spurious multimodal switching across chunk boundaries.
In a representative bowl-stacking rollout, RTC alternates (arrow) between competing grasp goals (green circle) and execution arms (red circle) over successive chunks, producing visibly hesitant corrections. Legato preserves a consistent grasp goal and arm choice (blue circle), leading to steadier progress.}
    \vspace{-3mm} 
    \label{fig:multimodal}
\end{figure*}

\subsubsection{\textbf{Tasks and Environments}}
\label{sec:tasks}
We evaluate our method on five real-world manipulation tasks:
(i) stack the bowls,
(ii) pour things into the bowl,
(iii) put all the items into the box,
(iv) fold the towel,
and (v) open the drawer, as shown in \cref{fig:env}.
These tasks jointly test different action patterns (e.g., rotation- or translation-dominant motions) and multimodal action selection (e.g., multiple valid grasp goals or the choice of different arms for execution). 
All tasks are evaluated with a fixed time cutoff of 120\,s.
Details are provided in the appendix.

\subsubsection{\textbf{Evaluation Metrics}}

\label{sec:metrics}
The following evaluation metrics are used to assess real-world experimental performance.

\textbf{Task completion score.}
Each rollout is assigned a task-specific completion score based on task progress and failure cases (e.g., partial success, object drops, or incorrect actions). Higher scores indicate better task completion.

\textbf{Task completion time.}
We measure the total time required to complete each task. This metric reflects the execution efficiency of the policy, capturing delays caused by hesitation or spurious action switching during real-world execution. 

\textbf{Trajectory smoothness metrics.}
Following prior work on action smoothness, we evaluate smoothness on the model output commands rather than robot executed states. This decouples model behavior from low-level controller performance.
Specifically, we report three smoothness-related metrics that capture complementary aspects of trajectory quality~\cite{Balasubramanian2015OnTA,arachchige2025sail}:
\begin{itemize}
    \item \textbf{Negative SPARC (NSPARC)}, where SPARC~\cite{arachchige2025sail,Balasubramanian2015OnTA} (Linear and Angular Spectral Arc Length) measures the smoothness of the velocity profile in the frequency domain over the \emph{entire trajectory}. Lower values of NSPARC indicate smoother global speed modulation with reduced high-frequency fluctuations.
    \item \textbf{Negative LDLJ (NLDLJ)}, where LDLJ~\cite{arachchige2025sail,Balasubramanian2015OnTA} (Log Dimensionless Jerk) quantifies high-order geometric smoothness by integrating squared jerk over the \emph{entire trajectory}. Lower NLDLJ correspond to reduced overall jerk energy and smoother motion at a global level.
    \item \textbf{Chunk-overlap RMSE}, computed over the overlapping delay segment between consecutive action chunks, which evaluates \emph{local trajectory continuity} at chunk connections rather than global smoothness.
\end{itemize}
Except for the task completion score, lower values indicate better performance for all metrics. Details of all metrics are provided in the appendix.

\subsubsection{\textbf{Models and Training Protocol}}
We compare the RTC baseline and our proposed Legato method under a strictly controlled setting. Both methods are initialized from the same $\pi_{0.5}$ pretrained checkpoint, trained on identical task datasets, and optimized using the same training hyperparameters and number of training steps.

\subsection{Main Results}

In this section, we report real-world evaluation results of Legato and RTC across five manipulation tasks executed on physical robotic platforms. As summarized in \cref{tab:real_world_main_results}, Legato consistently outperforms RTC across all evaluated tasks.

\subsubsection{\textbf{Task efficiency}}
Legato consistently achieves shorter task completion time than RTC across all tasks.
As analyzed in \cref{sec:native_continuation}, the schedule-shaped velocity reweighting increases the difficulty of switching between competing action modes, effectively suppressing frequent multimodal oscillations during execution.

Empirically, this leads to more decisive action generation with reduced hesitation before execution, thereby shortening overall task duration.
The effect is particularly pronounced in the bowl-stacking task, where multiple visually similar bowls induce a large number of plausible action modes, as shown in \cref{fig:multimodal}.
In such settings, RTC often alternates between competing strategies, while Legato maintains consistent mode selection and completes the task more efficiently.

\subsubsection{\textbf{Trajectory smoothness}}
Legato also demonstrates clear advantages in trajectory smoothness compared to RTC. With the exception of NLDLJ, all smoothness-related metrics show statistically significant improvements in favor of Legato.

Specifically, Legato consistently achieves lower NSPARC values across all tasks. This result indicates that Legato produces commands with reduced high-frequency velocity fluctuations and more regular speed modulation. Such improvements correspond to smoother and more visually coherent motions observed during real-world execution, as shown in the trajectories in \cref{fig:teaser}.

In addition, Legato substantially reduces the chunk-overlap RMSE across tasks. The observed improvements indicate that Legato generates more coherent chunk-to-chunk transitions, leading to improved continuity at action boundaries and smoother chunk-to-chunk stitching.

In contrast, improvements in NLDLJ do not consistently reach statistical significance across all tasks. NLDLJ measures high-order geometric smoothness by integrating squared jerk over the entire trajectory and is therefore dominated by motion segments outside the chunk overlap regions. Importantly, NLDLJ does not degrade under Legato compared to RTC, indicating that while trajectory continuity is improved at chunk boundaries, the remaining portions of the trajectory do not exhibit degraded smoothness.

\subsubsection{\textbf{Task success}}
Finally, Legato exceeds the task completion scores achieved by RTC. This confirms that the observed improvements in execution efficiency and trajectory smoothness do not come at the cost of task success, but instead translate into more reliable and effective real-world manipulation performance.


\begin{table}[t]
\centering
\caption{Comparison of Training-Time RTC and Legato. The guidance configuration of Legato is $d{=}8$, $s{=}30$, $r{=}22$.
Values are reported as mean $\pm$ standard error.}
\label{tab:ablation_train_rtc}

\small
\renewcommand{\arraystretch}{0.9}
\setlength{\tabcolsep}{6pt}

\begin{tabular}{l|cc}
\toprule
\textbf{Metric} 
& Training-time RTC
& \cellcolor{legatoblue}{Legato} \\
\midrule

Score $\uparrow$
& \pmstd{9.46}{0.16}
& \cellcolor{legatoblue}\textbf{\pmstd{9.72}{0.13}} \\

Completion Time (s) $\downarrow$
& \pmstd{81.73}{1.12}
& \cellcolor{legatoblue}\textbf{\pmstd{75.73}{1.51}} \\

NSPARC $\downarrow$
& \pmstd{2.46}{0.14}
& \cellcolor{legatoblue}\textbf{\pmstd{1.65}{0.08}} \\

NLDLJ $\downarrow$
& \pmstd{39.95}{0.13}
& \cellcolor{legatoblue}\textbf{\pmstd{39.50}{0.13}} \\

\bottomrule
\end{tabular}
\vspace{-3mm} 
\end{table}




\subsection{Comparison with Training-Time RTC}

We compare Legato with the recently proposed training-time RTC ~\cite{black2025training} on the \textit{pour} task, which also introduces continuation during training by constraining overlapping action segments.
We implement training-time RTC following the original formulation and compare it against Legato under the same experimental settings.
As shown in \cref{tab:ablation_train_rtc}, Legato achieves higher task scores, shorter completion times, and improved smoothness metrics compared to training-time RTC.

To contextualize this comparison, when the ramp length in our guidance schedule is set to zero, the schedule reduces to a hard overlap constraint that is similar in form to training-time RTC.
However, this similarity is limited to the constraint shape: the two approaches differ fundamentally in how continuation is incorporated into the learned policy.
Training-time RTC treats continuation as an external constraint via hard prefix conditioning while leaving the underlying flow dynamics unchanged.
In contrast, Legato reshapes the learned flow dynamics to match the effective denoising behavior induced by repeated, schedule-shaped guidance, so continuation becomes a native property of the policy dynamics.

Overall, these results suggest that reshaping the policy dynamics (rather than enforcing hard overlap constraints alone) is important for effective chunk continuation, and that using a non-zero ramp further enables smoother transitions between consecutive action chunks.

\begin{figure}[t]
    \centering
    \includegraphics[width=0.9\linewidth]{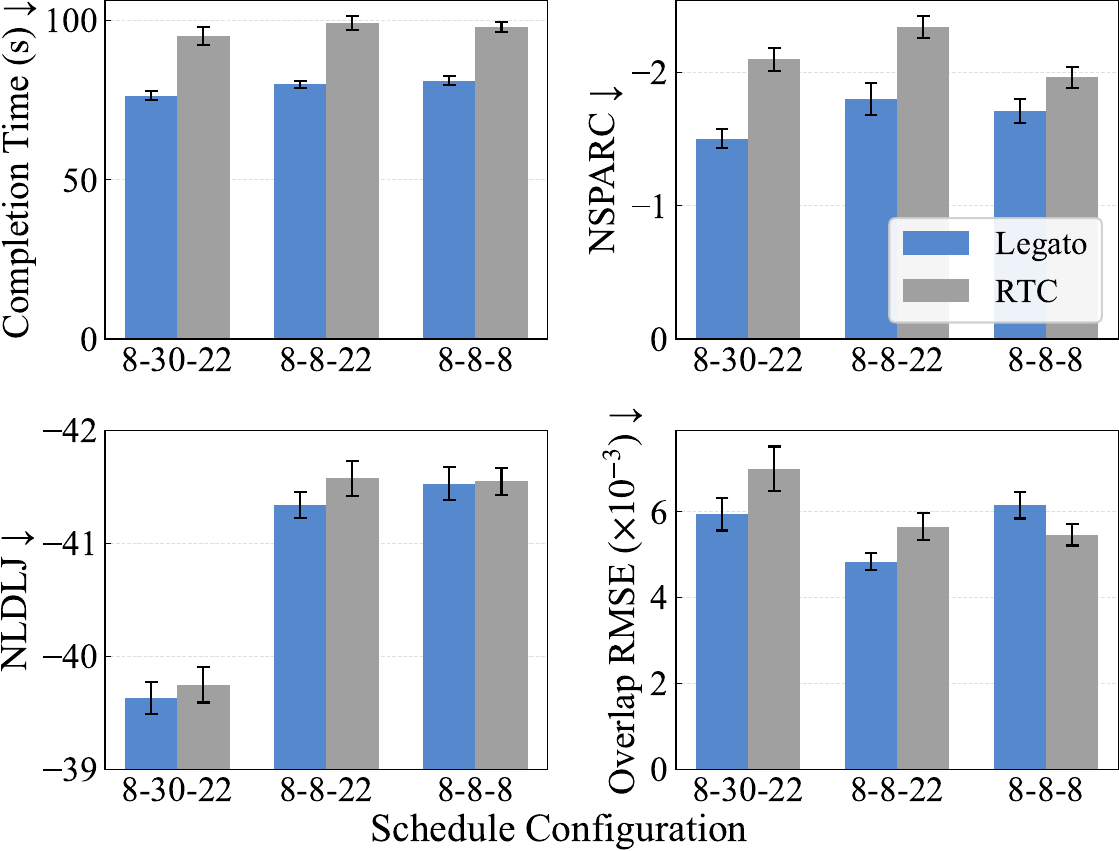}
    \caption{Schedule ablation reveals a controllable trade-off between local overlap consistency and smoothness.
Across schedule configurations ($d,s,r$), Legato outperforms RTC on completion time and smoothness. Decreasing stride strengthens overlap coupling but can increase high-frequency content; shortening the ramp partially recovers frequency-domain smoothness at the cost of weaker overlap alignment.}
    \label{fig:schedule-s}
    \vspace{-3mm} 
\end{figure}

\subsection{Ablation Studies}

In this section, we conduct a comprehensive set of ablation studies to analyze the applicability and robustness of the proposed method. Specifically, we examine: (i) the effect of different guidance schedule settings at inference time, (ii) the role of the condition row used in our policy, and (iii) the performance of Legato across different VLA models.

\subsubsection{\textbf{Varying the execution stride s}}
RTC recommends setting the execution stride $s$ to at least half of the action chunk length. However, since $s$ directly determines the effective inference frequency, a larger stride inevitably reduces the model’s responsiveness. This reveals an inherent trade-off between inference efficiency and control reactivity, motivating a detailed ablation over guidance schedule configurations.

When the execution stride $s$ becomes smaller than half of the chunk length, the ramp segment may extend beyond the immediate next chunk. To avoid that, we shorten the ramp segment as $s$ decreases, ensuring that the ramp always remains confined within the next chunk. We evaluate several guidance schedule configurations on the \textit{pour} task, as illustrated in \cref{fig:schedule-s}.

Our findings can be summarized as follows:




\paragraph{\textbf{Legato consistently outperforms RTC on almost all metrics}}
    The only exception is the overlap RMSE in the $d=s=r=8$ setting, which is discussed in the appendix.

\paragraph{\textbf{Reducing the execution stride $s$ improves chunk-to-chunk consistency but can degrade global smoothness}} Under the constraint $r + s + d = H$, a smaller $s$ implies a larger ramp length $r$, which improves chunk-to-chunk continuity, as reflected by lower overlap RMSE. At the same time, smaller strides lead to more frequent overlap regions, causing high-frequency components to accumulate and resulting in degraded whole-trajectory smoothness metrics.

\paragraph{\textbf{Shortening the ramp while keeping $s$ small improves frequency-domain smoothness at the expense of overlap consistency}} When $s$ remains small but the ramp length is shortened, NSPARC improves, indicating smoother frequency-domain behavior. However, this reduces overlap consistency, reflecting a weaker coupling between adjacent chunks.

Overall, these results demonstrate that the execution stride $s$ and ramp length jointly control a fundamental trade-off between local chunk connection quality and global frequency-domain smoothness. By adjusting their relative proportions, Legato enables flexible control over trajectory smoothness.


\begin{table}[t]
\centering
\caption{Ablation study on robustness to inference delay.
We vary the inference delay $d$ with a fixed execution stride $s$, where the ramp length $r$ changes accordingly due to the schedule constraint.
Values are reported as mean $\pm$ standard error.}
\label{tab:ablation_delay}

\small
\renewcommand{\arraystretch}{0.9}
\setlength{\tabcolsep}{8pt}

\begin{tabular}{cc|cc}
\toprule
\textbf{$(d,s,r)$} 
& \textbf{Method}
& \textbf{NSPARC} $\downarrow$
& \textbf{Overlap RMSE} $\downarrow$ \\
\midrule

\multirow{2}{*}{(10,30,20)}
& RTC
& \pmstd{2.03}{0.08}
& \pmstd{9.23}{0.75} \\
& \cellcolor{legatoblue}{Legato}
& \cellcolor{legatoblue}\textbf{\pmstd{1.68}{0.09}}
& \cellcolor{legatoblue}\textbf{\pmstd{7.00}{0.50}} \\
\midrule

\multirow{2}{*}{(8,30,22)}
& RTC
& \pmstd{2.10}{0.09}
& \pmstd{7.00}{0.52} \\
& \cellcolor{legatoblue}{Legato}
& \cellcolor{legatoblue}\textbf{\pmstd{1.50}{0.07}}
& \cellcolor{legatoblue}\textbf{\pmstd{5.94}{0.38}} \\
\midrule

\multirow{2}{*}{(6,30,24)}
& RTC
& \pmstd{2.03}{0.08}
& \pmstd{9.23}{0.75} \\
& \cellcolor{legatoblue}{Legato}
& \cellcolor{legatoblue}\textbf{\pmstd{1.38}{0.05}}
& \cellcolor{legatoblue}\textbf{\pmstd{5.44}{0.31}} \\
\bottomrule
\end{tabular}
\vspace{-3mm}
\end{table}

\subsubsection{\textbf{Varying the inference delay d}}
In addition to the execution stride, the inference delay $d$ also plays an important role in shaping trajectory smoothness. To isolate its effect, we fix the execution stride $s$ and vary the delay length $d$, conducting evaluations on the \textit{pour} task. The quantitative results are summarized in \cref{tab:ablation_delay}.

Across all evaluated metrics, Legato consistently outperforms RTC, demonstrating the robustness of the proposed method to variations in inference latency. When analyzing Legato specifically, we find that reducing the delay length decreases the size of the overlap region while simultaneously increasing the relative length of the ramp segment. This leads to improved chunk-to-chunk continuity and smoother execution, as reflected by better overlap consistency and frequency-domain smoothness metrics.

Overall, these results indicate that both execution stride $s$ and inference delay $d$ provide effective control knobs for shaping smoothness properties of generated trajectories. Legato can flexibly adapt to different schedule configurations while consistently maintaining superior performance over RTC.


\begin{table}[t]
\centering
\caption{Ablation study on the effect of the condition row under different guidance configurations.
We vary the inference delay $d$ and ramp length $r$ to construct different schedules.
Values are reported as mean $\pm$ standard error.}
\label{tab:ablation_cond_row}

\small
\renewcommand{\arraystretch}{0.9}
\setlength{\tabcolsep}{8pt}

\begin{tabular}{cc|cc}
\toprule
\textbf{$(d,s,r)$} 
& \textbf{Method}
& \textbf{NSPARC} $\downarrow$
& \textbf{Overlap RMSE} $\downarrow$ \\
\midrule

\multirow{2}{*}{(10,30,20)}
& w/o cond
& \textbf{\pmstd{1.64}{0.07}}
& \pmstd{7.88}{0.70} \\
& \cellcolor{legatoblue}{w/ cond}
& \cellcolor{legatoblue}\pmstd{1.68}{0.09}
& \cellcolor{legatoblue}\textbf{\pmstd{7.00}{0.50}} \\
\midrule

\multirow{2}{*}{(8,30,22)}
& w/o cond
& \pmstd{1.52}{0.09}
& \pmstd{7.21}{0.68} \\
& \cellcolor{legatoblue}{w/ cond}
& \cellcolor{legatoblue}\textbf{\pmstd{1.50}{0.07}}
& \cellcolor{legatoblue}\textbf{\pmstd{5.94}{0.38}} \\
\midrule

\multirow{2}{*}{(6,30,24)}
& w/o cond
& \pmstd{1.49}{0.10}
& \pmstd{6.40}{0.52} \\
& \cellcolor{legatoblue}{w/ cond}
& \cellcolor{legatoblue}\textbf{\pmstd{1.38}{0.05}}
& \cellcolor{legatoblue}\textbf{\pmstd{5.44}{0.31}} \\

\bottomrule
\end{tabular}
\end{table}












\begin{table}[t]
\centering
\caption{Ablation results on the $\pi_0$ and Gr00t N1.6 model comparing RTC and Legato under the same guidance configuration ($d{=}8$, $s{=}30$, $r{=}22$).
Values are reported as mean $\pm$ standard error.}
\label{tab:ablation_diffmodel}

\small
\renewcommand{\arraystretch}{0.9}
\setlength{\tabcolsep}{10pt}

\begin{tabular}{l|cc}
\toprule
\textbf{Metric}
& $\pi_0$ + RTC
& \cellcolor{legatoblue}{$\pi_0$ + Legato} \\
\midrule

Completion Time $\downarrow$
& \pmstd{92.93}{1.90}
& \cellcolor{legatoblue}\textbf{\pmstd{88.30}{1.29}} \\

NSPARC $\downarrow$
& \pmstd{2.00}{0.09}
& \cellcolor{legatoblue}\textbf{\pmstd{1.83}{0.08}} \\

NLDLJ $\downarrow$
& \pmstd{40.48}{0.21}
& \cellcolor{legatoblue}\textbf{\pmstd{40.27}{0.09}} \\

Overlap RMSE $\downarrow$
& \pmstd{8.63}{0.65}
& \cellcolor{legatoblue}\textbf{\pmstd{7.50}{0.49}} \\

\midrule

\textbf{Metric}
& Gr00t + RTC
& \cellcolor{legatoblue}{Gr00t + Legato} \\
\midrule

Completion Time $\downarrow$
& \pmstd{91.60}{2.30}
& \cellcolor{legatoblue}\textbf{\pmstd{84.90}{1.30}} \\

NSPARC $\downarrow$
& \pmstd{2.99}{0.17}
& \cellcolor{legatoblue}\textbf{\pmstd{2.63}{0.06}} \\

NLDLJ $\downarrow$
& \pmstd{39.91}{0.07}
& \cellcolor{legatoblue}\textbf{\pmstd{39.62}{0.12}} \\

Overlap RMSE $\downarrow$
& \pmstd{9.43}{0.46}
& \cellcolor{legatoblue}\textbf{\pmstd{8.75}{0.36}} \\

\bottomrule
\end{tabular}

\vspace{-2mm}
\end{table}

\subsubsection{\textbf{Condition Row}}


To evaluate whether the condition row is useful, we conduct an ablation study in which the guidance schedule is no longer provided as an explicit condition.

We perform this ablation on the \textit{pour} task, and report the results in \cref{tab:ablation_cond_row}. As shown, removing the condition row leads to a degradation in performance, particularly in trajectory smoothness and execution stability. This suggests that explicitly providing the guidance schedule helps the model disambiguate different continuation regimes induced by varying $(d, r)$ pairs, and enables more reliable adaptation to dynamic inference conditions.


\subsubsection{\textbf{Different Models}}
In the main results \cref{tab:real_world_main_results}, we evaluate our method on the $\pi_{0.5}$ model. To evaluate whether the proposed method generalizes across different VLA models, we further conduct experiments on the $\pi_0$ and Gr00t N1.6 model. We select the representative task \textit{pour things}.
As shown in \cref{tab:ablation_diffmodel}, Legato outperforms RTC on the $\pi_0$ and Gr00t N1.6 model on the pour task.

These results demonstrate that the proposed method is not tied to a specific policy backbone or training configuration, and can be effectively transferred across different flow-based VLA models, highlighting its robustness and model generality.

\subsubsection{\textbf{Different Smoothing Methods}}



\vspace{2pt}
\begin{table}[t]
\centering
\caption{Comparison of TE and Legato on the pour task.}
\label{tab:te_legato}
\small
\setlength{\tabcolsep}{8pt}
\begin{tabular}{lccc}
\toprule
\textbf{Method}
& \textbf{Score}$\uparrow$
& \textbf{Time (s)}$\downarrow$
& \textbf{NSPARC}$\downarrow$ \\
\midrule
TE
& \pmstd{9.17}{0.16}
& \pmstd{96.87}{2.79}
& \pmstd{2.96}{0.23} \\
\cellcolor{legatoblue}{Legato}
& \cellcolor{legatoblue}\textbf{\pmstd{9.67}{0.10}}
& \cellcolor{legatoblue}\textbf{\pmstd{76.43}{1.46}}
& \cellcolor{legatoblue}\textbf{\pmstd{1.50}{0.07}} \\
\bottomrule
\end{tabular}


\end{table}

As diffusion-based methods are not directly comparable, we compare with the trajectory smoothing baseline Temporal Ensembling (TE)~\cite{zhao2023learning}. This method maintains a buffer of predicted chunks and executes averaged actions for each timestep.
As shown in \cref{tab:te_legato}, Legato outperforms TE in both execution time and continuation quality, benefiting from its native continuation mechanism.

\subsubsection{\textbf{Sensitivity to denoising steps}}
\label{sec:denoising}


Training velocity implicitly depends on the denoising step count $N$, while we additionally evaluate generalization under different inference-time denoising schedules.
To improve robustness across varying $N$, we apply a heuristic rescaling
$\omega'_i{=}\text{clamp}(1{-}(1{-}\omega_i){\cdot}(N_\text{train}/N_\text{infer}), 0, 1)$
at inference time.
Although this rescaling introduces a mild training-inference mismatch, since the velocity field is trained under a fixed $N_\text{train}$, we find that it substantially reduces sensitivity to the inference denoising step count in practice.
Without the condition row, Legato even remains effective under single-step denoising, as shown in \cref{tab:denoise_step}.

\begin{table}[t]
\centering
\caption{NSPARC metric results under varying inference $N$ (trained for $N{=}5$ on pour task) of Legato.}
\label{tab:denoise_step}
{\scriptsize
\setlength{\tabcolsep}{4pt}
\begin{tabular}{lcccc}
\toprule
 & $N{=}1$ & $N{=}3$ & $N{=}10$ & $N{=}15$\\
\midrule
w/o cond.\ row & $1.98 \pm 0.17$ & $1.85 \pm 0.07$ & ${1.77 \pm 0.18}$ & $1.97 \pm 0.09$ \\
w/ cond.\ row  & \textit{unstable motion} & $2.65 \pm 0.11$ & $1.82 \pm 0.16$ & $2.11 \pm 0.13$ \\
\bottomrule
\end{tabular}}

\end{table}

\subsection{\textbf{Simulation Results on Kinetix}}
\begin{figure}[t]
\centering
\includegraphics[width=1.0\linewidth]{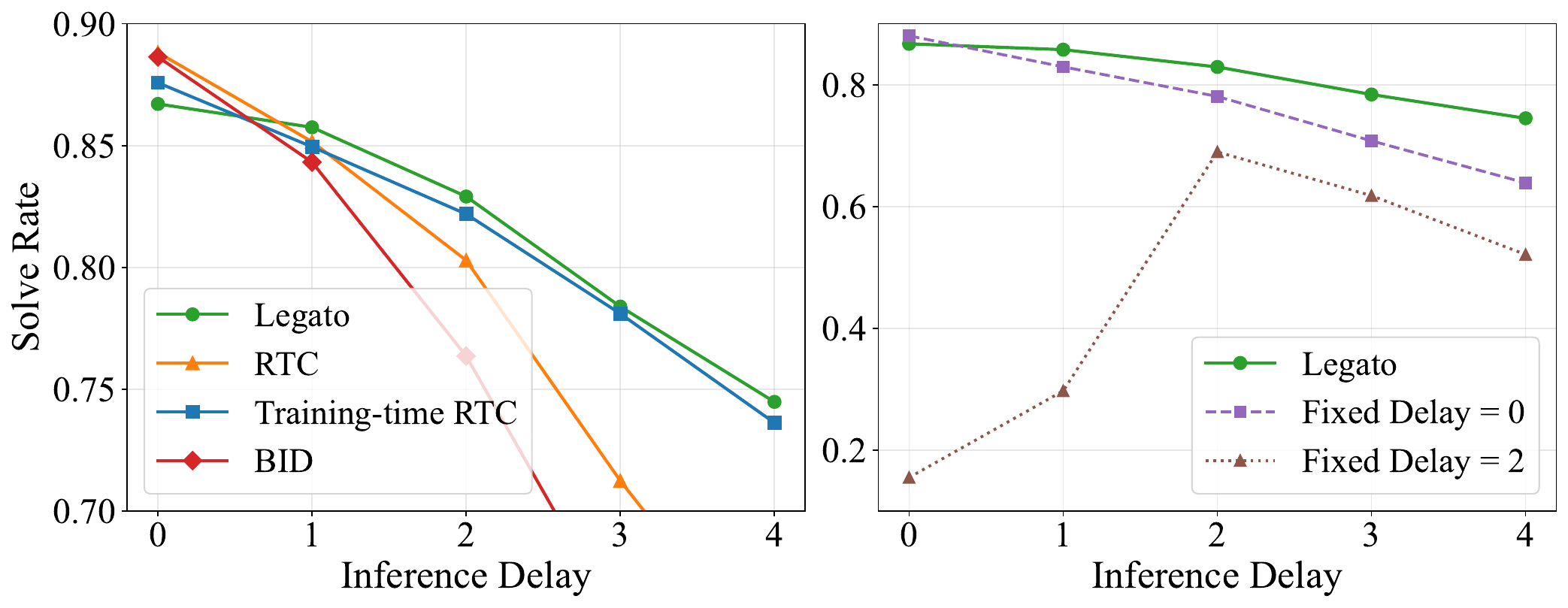}
\vspace{-6pt}
\caption{Left: Legato vs.\ RTC, Training-time RTC, and BID in Kinetix. Right: Ablation result on the schedule randomization.}
\label{fig:kinetix}
\vspace{-6pt}
\end{figure}

\label{sec:simulation}
We test Legato in Kinetix, using identical settings to Training-time RTC.
As shown in \Cref{fig:kinetix} (left): Legato outperforms all methods when delay${>}0$. In Kinetix, the policy is a based on a UNet, shown the generalization of Legato across different flow-based policies.

We also ablate schedule randomization during training in Kinetix.
As shown in \Cref{fig:kinetix} (right), fixing $d$ during training degrades performance at other $d$ values during evaluation, highlighting the importance of delay randomization.
Notably, when the randomized delay range does not include $d{=}0$, the policy becomes severely out-of-distribution at the first denoising step without guidance, leading to near-complete failure during evaluation.

\subsection{\textbf{Deployment guidelines}}

We recommend the following hyper-parameters for deployment: $d{=}\max(\text{recent delays})$, $s{=}0.5H$, $r{=}H{-}d{-}s$.
In practice, following RTC, we maintain a short history of recent inference delays and use the maximum observed delay as the deployment-time $d$.
For variable-$N$ inference, we recommend using the w/o-condition-row variant.

Additionally, we find that finetuning from a strong base policy leads to more stable convergence and reduces the tendency of excessive guidance training to impair reactivity. Therefore, we recommend initializing Legato from a strong pretrained base policy.
\section{Conclusion}

In this work, we propose Legato, a training-time continuation method for action-chunked flow-based VLA policies. Legato reshapes the learned flow dynamics to align training and inference under schedule-shaped, per-step continuation, making chunk continuation a native property of the policy.
This design improves trajectory smoothness and reduces spurious multimodal switching at chunk boundaries, leading to smoother action and more consistent action modes, less hesitation, and shorter task completion time. By conditioning on randomized schedules, a single policy can adapt to different inference delays and flexibly control trajectory smoothness.

In the current formulation, the denoise step is specified at training time, limiting the ability to adjust it during inference.
Future work could investigate more flexible native continuation schemes with consistent training and inference dynamics.

\section*{Acknowledgments}
This research was conducted with the support of the Shanghai Qi Zhi Institute \& Spirit AI Innovation Program and the Tsinghua University Dushi Program. Funding and support for this work were also provided by the Tsinghua University - Keystone Electrical (Zhejiang) Co.,Ltd Joint Research Center for Embodied Multimodal Artificial Intelligence (JCEMAI). Additionally, we would like to extend our thanks to the Xiongan AI Institute.

\balance
\bibliographystyle{plainnat}
\bibliography{references}

@article{black2024pi_0,
  title={$\pi_0$: A Vision-Language-Action Flow Model for General Robot Control},
  author={Black, Kevin and Brown, Noah and Driess, Danny and Esmail, Adnan and Equi, Michael and Finn, Chelsea and Fusai, Niccolo and Groom, Lachy and Hausman, Karol and Ichter, Brian and others},
  journal={arXiv preprint arXiv:2410.24164},
  year={2024}
}

@inproceedings{black2025pi_,
  title={$\pi_{0.5}$: a Vision-Language-Action Model with Open-World Generalization},
  author={Black, Kevin and Brown, Noah and Darpinian, James and Dhabalia, Karan and Driess, Danny and Esmail, Adnan and Equi, Michael Robert and Finn, Chelsea and Fusai, Niccolo and Galliker, Manuel Y and others},
  booktitle={9th Annual Conference on Robot Learning},
  year={2025}
}

@article{chi2025diffusion,
  title={Diffusion policy: Visuomotor policy learning via action diffusion},
  author={Chi, Cheng and Xu, Zhenjia and Feng, Siyuan and Cousineau, Eric and Du, Yilun and Burchfiel, Benjamin and Tedrake, Russ and Song, Shuran},
  journal={The International Journal of Robotics Research},
  volume={44},
  number={10-11},
  pages={1684--1704},
  year={2025},
  publisher={Sage Publications Sage UK: London, England}
}

@article{pertsch2025fast,
  title={Fast: Efficient action tokenization for vision-language-action models},
  author={Pertsch, Karl and Stachowicz, Kyle and Ichter, Brian and Driess, Danny and Nair, Suraj and Vuong, Quan and Mees, Oier and Finn, Chelsea and Levine, Sergey},
  journal={arXiv preprint arXiv:2501.09747},
  year={2025}
}

@article{liu2024bidirectional,
  title={Bidirectional decoding: Improving action chunking via closed-loop resampling},
  author={Liu, Yuejiang and Hamid, Jubayer Ibn and Xie, Annie and Lee, Yoonho and Du, Maximilian and Finn, Chelsea},
  journal={arXiv preprint arXiv:2408.17355},
  year={2024}
}

@article{black2025real,
  title={Real-Time Execution of Action Chunking Flow Policies},
  author={Black, Kevin and Galliker, Manuel Y and Levine, Sergey},
  journal={arXiv preprint arXiv:2506.07339},
  year={2025}
}

@article{chen2024diffusion,
  title={Diffusion forcing: Next-token prediction meets full-sequence diffusion},
  author={Chen, Boyuan and Mart{\'\i} Mons{\'o}, Diego and Du, Yilun and Simchowitz, Max and Tedrake, Russ and Sitzmann, Vincent},
  journal={Advances in Neural Information Processing Systems},
  volume={37},
  pages={24081--24125},
  year={2024}
}

@article{hoeg2024streaming,
  title={Streaming diffusion policy: Fast policy synthesis with variable noise diffusion models},
  author={H{\o}eg, Sigmund H and Du, Yilun and Egeland, Olav},
  journal={arXiv preprint arXiv:2406.04806},
  year={2024}
}

@inproceedings{jung2025rolling,
  title={Rolling Diffusion Policy for Robotic Action Prediction: Enhancing Efficiency and Temporal Awareness},
  author={Jung, Chanhyuk and Ahn, Dasom and Kim, Sangwon and Jang, In-su and Kim, Kwang-Ju and Yoo, Sungkeun and Ko, Byoung Chul},
  booktitle={ICRA 2025 Workshop on Foundation Models and Neuro-Symbolic AI for Robotics},
  year={2025}
}

@article{arachchige2025sail,
  title={SAIL: Faster-than-Demonstration Execution of Imitation Learning Policies},
  author={Arachchige, Nadun Ranawaka and Chen, Zhenyang and Jung, Wonsuhk and Shin, Woo Chul and Bansal, Rohan and Barroso, Pierre and He, Yu Hang and Lin, Yingyang Celine and Joffe, Benjamin and Kousik, Shreyas and others},
  journal={arXiv preprint arXiv:2506.11948},
  year={2025}
}

@article{black2025training,
  title={Training-Time Action Conditioning for Efficient Real-Time Chunking},
  author={Black, Kevin and Ren, Allen Z and Equi, Michael and Levine, Sergey},
  journal={arXiv preprint arXiv:2512.05964},
  year={2025}
}

@inproceedings{zitkovich2023rt,
  title={Rt-2: Vision-language-action models transfer web knowledge to robotic control},
  author={Zitkovich, Brianna and Yu, Tianhe and Xu, Sichun and Xu, Peng and Xiao, Ted and Xia, Fei and Wu, Jialin and Wohlhart, Paul and Welker, Stefan and Wahid, Ayzaan and others},
  booktitle={Conference on Robot Learning},
  pages={2165--2183},
  year={2023},
  organization={PMLR}
}

@article{kim2024openvla,
  title={Openvla: An open-source vision-language-action model},
  author={Kim, Moo Jin and Pertsch, Karl and Karamcheti, Siddharth and Xiao, Ted and Balakrishna, Ashwin and Nair, Suraj and Rafailov, Rafael and Foster, Ethan and Lam, Grace and Sanketi, Pannag and others},
  journal={arXiv preprint arXiv:2406.09246},
  year={2024}
}

@article{wen2025tinyvla,
  title={Tinyvla: Towards fast, data-efficient vision-language-action models for robotic manipulation},
  author={Wen, Junjie and Zhu, Yichen and Li, Jinming and Zhu, Minjie and Tang, Zhibin and Wu, Kun and Xu, Zhiyuan and Liu, Ning and Cheng, Ran and Shen, Chaomin and others},
  journal={IEEE Robotics and Automation Letters},
  year={2025},
  publisher={IEEE}
}

@article{lipman2022flow,
  title={Flow matching for generative modeling},
  author={Lipman, Yaron and Chen, Ricky TQ and Ben-Hamu, Heli and Nickel, Maximilian and Le, Matt},
  journal={arXiv preprint arXiv:2210.02747},
  year={2022}
}

@article{zhao2023learning,
  title={Learning fine-grained bimanual manipulation with low-cost hardware},
  author={Zhao, Tony Z and Kumar, Vikash and Levine, Sergey and Finn, Chelsea},
  journal={arXiv preprint arXiv:2304.13705},
  year={2023}
}

@article{bjorck2025gr00t,
  title={Gr00t n1: An open foundation model for generalist humanoid robots},
  author={Bjorck, Johan and Casta{\~n}eda, Fernando and Cherniadev, Nikita and Da, Xingye and Ding, Runyu and Fan, Linxi and Fang, Yu and Fox, Dieter and Hu, Fengyuan and Huang, Spencer and others},
  journal={arXiv preprint arXiv:2503.14734},
  year={2025}
}

@article{liu2024rdt,
  title={Rdt-1b: a diffusion foundation model for bimanual manipulation},
  author={Liu, Songming and Wu, Lingxuan and Li, Bangguo and Tan, Hengkai and Chen, Huayu and Wang, Zhengyi and Xu, Ke and Su, Hang and Zhu, Jun},
  journal={arXiv preprint arXiv:2410.07864},
  year={2024}
}

@article{Lin2025OneTwoVLAAU,
  title={OneTwoVLA: A Unified Vision-Language-Action Model with Adaptive Reasoning},
  author={Fanqi Lin and Ruiqian Nai and Yingdong Hu and Jiacheng You and Junming Zhao and Yang Gao},
  journal={ArXiv},
  year={2025},
  volume={abs/2505.11917}
}

@article{team2024octo,
  title={Octo: An open-source generalist robot policy},
  author={Team, Octo Model and Ghosh, Dibya and Walke, Homer and Pertsch, Karl and Black, Kevin and Mees, Oier and Dasari, Sudeep and Hejna, Joey and Kreiman, Tobias and Xu, Charles and others},
  journal={arXiv preprint arXiv:2405.12213},
  year={2024}
}

@article{Zhao2025CoTVLAVC,
  title={CoT-VLA: Visual Chain-of-Thought Reasoning for Vision-Language-Action Models},
  author={Qingqing Zhao and Yao Lu and Moo Jin Kim and Zipeng Fu and Zhuoyang Zhang and Yecheng Wu and Zhaoshuo Li and Qianli Ma and Song Han and Chelsea Finn and Ankur Handa and Ming-Yu Liu and Donglai Xiang and Gordon Wetzstein and Tsung-Yi Lin},
  journal={2025 IEEE/CVF Conference on Computer Vision and Pattern Recognition (CVPR)},
  year={2025},
  pages={1702-1713}
}

@article{Wang2025VQVLAIV,
  title={VQ-VLA: Improving Vision-Language-Action Models via Scaling Vector-Quantized Action Tokenizers},
  author={Yating Wang and Haoyi Zhu and Mingyu Liu and Jiange Yang and Hao-Shu Fang and Tong He},
  journal={ArXiv},
  year={2025},
  volume={abs/2507.01016}
}

@article{qu2025eo,
  title={Eo-1: Interleaved vision-text-action pretraining for general robot control},
  author={Qu, Delin and Song, Haoming and Chen, Qizhi and Chen, Zhaoqing and Gao, Xianqiang and Ye, Xinyi and Lv, Qi and Shi, Modi and Ren, Guanghui and Ruan, Cheng and others},
  journal={arXiv preprint arXiv:2508.21112},
  year={2025}
}

@article{liang2025discrete,
  title={Discrete diffusion vla: Bringing discrete diffusion to action decoding in vision-language-action policies},
  author={Liang, Zhixuan and Li, Yizhuo and Yang, Tianshuo and Wu, Chengyue and Mao, Sitong and Nian, Tian and Pei, Liuao and Zhou, Shunbo and Yang, Xiaokang and Pang, Jiangmiao and others},
  journal={arXiv preprint arXiv:2508.20072},
  year={2025}
}

@article{wen2025dvla,
  title={dvla: Diffusion vision-language-action model with multimodal chain-of-thought},
  author={Wen, Junjie and Zhu, Minjie and Liu, Jiaming and Liu, Zhiyuan and Yang, Yicun and Zhang, Linfeng and Zhang, Shanghang and Zhu, Yichen and Xu, Yi},
  journal={arXiv preprint arXiv:2509.25681},
  year={2025}
}

@article{wen2025llada,
  title={Llada-vla: Vision language diffusion action models},
  author={Wen, Yuqing and Li, Hebei and Gu, Kefan and Zhao, Yucheng and Wang, Tiancai and Sun, Xiaoyan},
  journal={arXiv preprint arXiv:2509.06932},
  year={2025}
}

@article{lin2025evo,
  title={Evo-1: Lightweight vision-language-action model with preserved semantic alignment},
  author={Lin, Tao and Zhong, Yilei and Du, Yuxin and Zhang, Jingjing and Liu, Jiting and Chen, Yinxinyu and Gu, Encheng and Liu, Ziyan and Cai, Hongyi and Zou, Yanwen and others},
  journal={arXiv preprint arXiv:2511.04555},
  year={2025}
}

@article{Pokle2023TrainingfreeLI,
  title={Training-free linear image inverses via flows},
  author={Ashwini Pokle and Matthew Muckley and Ricky T. Q. Chen and Brian Karrer},
  journal={Trans. Mach. Learn. Res.},
  year={2023},
  volume={2024}
}

@inproceedings{Song2023PseudoinverseGuidedDM,
  title={Pseudoinverse-Guided Diffusion Models for Inverse Problems},
  author={Jiaming Song and Arash Vahdat and Morteza Mardani and Jan Kautz},
  booktitle={International Conference on Learning Representations},
  year={2023}
}

@article{Balasubramanian2015OnTA,
  title={On the analysis of movement smoothness},
  author={Sivakumar Balasubramanian and Alejandro Melendez-Calderon and Agn{\`e}s Roby-Brami and Etienne Burdet},
  journal={Journal of NeuroEngineering and Rehabilitation},
  year={2015},
  volume={12}
}

@article{pearce2023imitating,
  title={Imitating human behaviour with diffusion models},
  author={Pearce, Tim and Rashid, Tabish and Kanervisto, Anssi and Bignell, Dave and Sun, Mingfei and Georgescu, Raluca and Macua, Sergio Valcarcel and Tan, Shan Zheng and Momennejad, Ida and Hofmann, Katja and others},
  journal={arXiv preprint arXiv:2301.10677},
  year={2023}
}

@article{lai2022action,
  title={Action chunking as policy compression},
  author={Lai, Lucy and Huang, Ann Zixiang and Gershman, Samuel J},
  journal={PsyArXiv},
  year={2022}
}

@article{team2025gemini,
  title={Gemini robotics: Bringing ai into the physical world},
  author={Team, Gemini Robotics and Abeyruwan, Saminda and Ainslie, Joshua and Alayrac, Jean-Baptiste and Arenas, Montserrat Gonzalez and Armstrong, Travis and Balakrishna, Ashwin and Baruch, Robert and Bauza, Maria and Blokzijl, Michiel and others},
  journal={arXiv preprint arXiv:2503.20020},
  year={2025}
}

@article{ho2020denoising,
  title={Denoising diffusion probabilistic models},
  author={Ho, Jonathan and Jain, Ajay and Abbeel, Pieter},
  journal={Advances in neural information processing systems},
  volume={33},
  pages={6840--6851},
  year={2020}
}

@article{chi2024universal,
  title={Universal manipulation interface: In-the-wild robot teaching without in-the-wild robots},
  author={Chi, Cheng and Xu, Zhenjia and Pan, Chuer and Cousineau, Eric and Burchfiel, Benjamin and Feng, Siyuan and Tedrake, Russ and Song, Shuran},
  journal={arXiv preprint arXiv:2402.10329},
  year={2024}
}

@article{cheang2025gr,
  title={Gr-3 technical report},
  author={Cheang, Chilam and Chen, Sijin and Cui, Zhongren and Hu, Yingdong and Huang, Liqun and Kong, Tao and Li, Hang and Li, Yifeng and Liu, Yuxiao and Ma, Xiao and others},
  journal={arXiv preprint arXiv:2507.15493},
  year={2025}
}

@article{zheng2024tracevla,
  title={Tracevla: Visual trace prompting enhances spatial-temporal awareness for generalist robotic policies},
  author={Zheng, Ruijie and Liang, Yongyuan and Huang, Shuaiyi and Gao, Jianfeng and Daum{\'e} III, Hal and Kolobov, Andrey and Huang, Furong and Yang, Jianwei},
  journal={arXiv preprint arXiv:2412.10345},
  year={2024}
}

@article{ho2022classifier,
  title={Classifier-free diffusion guidance},
  author={Ho, Jonathan and Salimans, Tim},
  journal={arXiv preprint arXiv:2207.12598},
  year={2022}
}

@article{zhen20243d,
  title={3d-vla: A 3d vision-language-action generative world model},
  author={Zhen, Haoyu and Qiu, Xiaowen and Chen, Peihao and Yang, Jincheng and Yan, Xin and Du, Yilun and Hong, Yining and Gan, Chuang},
  journal={arXiv preprint arXiv:2403.09631},
  year={2024}
}

@article{barreiros2025careful,
  title={A careful examination of large behavior models for multitask dexterous manipulation},
  author={Barreiros, Jose and Beaulieu, Andrew and Bhat, Aditya and Cory, Rick and Cousineau, Eric and Dai, Hongkai and Fang, Ching-Hsin and Hashimoto, Kunimatsu and Irshad, Muhammad Zubair and Itkina, Masha and others},
  journal={arXiv preprint arXiv:2507.05331},
  year={2025}
}

@inproceedings{braun2024riemannian,
  title={Riemannian flow matching policy for robot motion learning},
  author={Braun, Max and Jaquier, No{\'e}mie and Rozo, Leonel and Asfour, Tamim},
  booktitle={2024 IEEE/RSJ International Conference on Intelligent Robots and Systems (IROS)},
  pages={5144--5151},
  year={2024},
  organization={IEEE}
}

@misc{belkhale2024minivla,
  title={Minivla: A better vla with a smaller footprint},
  author={Belkhale, Suneel and Sadigh, Dorsa},
  year={2024},
  publisher={Stanford-ILIAD GitHub: openvla-mini}
}

@article{shukor2025smolvla,
  title={Smolvla: A vision-language-action model for affordable and efficient robotics},
  author={Shukor, Mustafa and Aubakirova, Dana and Capuano, Francesco and Kooijmans, Pepijn and Palma, Steven and Zouitine, Adil and Aractingi, Michel and Pascal, Caroline and Russi, Martino and Marafioti, Andres and others},
  journal={arXiv preprint arXiv:2506.01844},
  year={2025}
}

@article{yu2026twinbrainvla,
  title={TwinBrainVLA: Unleashing the Potential of Generalist VLMs for Embodied Tasks via Asymmetric Mixture-of-Transformers},
  author={Yu, Bin and Lian, Shijie and Lin, Xiaopeng and Wei, Yuliang and Shen, Zhaolong and Wu, Changti and Miao, Yuzhuo and Wang, Xinming and Wang, Bailing and Huang, Cong and others},
  journal={arXiv preprint arXiv:2601.14133},
  year={2026}
}

@article{yu2025point,
  title={Point What You Mean: Visually Grounded Instruction Policy},
  author={Yu, Hang and Zhao, Juntu and Liu, Yufeng and Li, Kaiyu and Ma, Cheng and Zhang, Di and Hu, Yingdong and Chen, Guang and Xie, Junyuan and Guo, Junliang and others},
  journal={arXiv preprint arXiv:2512.18933},
  year={2025}
}

@article{zhao2025you,
  title={Do You Need Proprioceptive States in Visuomotor Policies?},
  author={Zhao, Juntu and Lu, Wenbo and Zhang, Di and Liu, Yufeng and Liang, Yushen and Zhang, Tianluo and Cao, Yifeng and Xie, Junyuan and Hu, Yingdong and Wang, Shengjie and others},
  journal={arXiv preprint arXiv:2509.18644},
  year={2025}
}

\clearpage
\appendices

\section*{Appendix}
\addcontentsline{toc}{section}{Appendix} 
\renewcommand{\theequation}{A.\arabic{equation}}
\setcounter{equation}{0}
\renewcommand{\thefigure}{A.\arabic{figure}}
\setcounter{figure}{0}
\renewcommand{\thetable}{A.\arabic{table}}
\setcounter{table}{0}

\subsection{\textbf{Task Details}}

We evaluate all methods on five real-world manipulation tasks that span diverse object interactions, action patterns, and execution characteristics. 
Across all tasks, the robot starts from an identical initial configuration for all models. 
Unless otherwise specified, object positions, orientations, and appearances are randomized per trial but kept identical across different models to ensure fair comparison.
Unless otherwise noted, for all tasks except bowl, each model is evaluated over 30 trials.
All ablation studies follow the same evaluation protocol, with 30 trials conducted per model for each task.

\subsubsection{\textbf{Bowl Stacking (bowl)}}
The objective of this task is to stack all bowls placed on a tabletop into a single vertical stack.
We consider five settings with the number of bowls equal to $\{3, 4, 5, 6, 7\}$.
For each setting, 10 trials are conducted, resulting in a total of 50 trials.
In each trial, the initial positions and colors of the bowls are randomly sampled.
To ensure fair comparison, the same set of 50 initial configurations is used across all models.
A trial is considered successful if all bowls are stacked into one pile without any bowl falling off the table.

\subsubsection{\textbf{Pouring (pour)}}
This task evaluates coordinated grasping, lifting, and rotational control.
Two bowls of different colors are placed on the tabletop, one of which initially contains a set of small blocks.
The robot is required to grasp the bowl containing the blocks, pour all blocks into the empty bowl, then grasp the second bowl and pour the blocks back into the original bowl.
This sequence constitutes one complete pouring operation, as illustrated in fig. 1.
Each trial consists of three consecutive pouring operations.

\subsubsection{\textbf{Pick-and-Place (pickplace)}}
In this task, the robot must place all items on the table into a white box.
The objects include a small jar, a marker pen, and a small ball.
The white box and all three objects are randomly placed on the tabletop at the beginning of each trial, with configurations shared across models.
A trial is considered successful if all three objects are fully placed inside the box.

\subsubsection{\textbf{Drawer Opening (drawer)}}
This task requires the robot to open the second drawer of a white three-layer drawer cabinet.
At the beginning of each trial, the drawer cabinet is placed on the table with a randomly sampled position and orientation, while remaining consistent across models.
The task is considered successful if the second drawer is pulled open beyond a predefined distance threshold.

\subsubsection{\textbf{Towel Folding (towel)}}
The objective of this task is to fold a towel placed on the tabletop.
The towel’s initial position and orientation are randomly sampled for each trial and kept identical across models.
A trial is considered successful if the towel is folded into a compact configuration according to predefined geometric criteria.

\subsection{\textbf{Delay Construction Details}}

In our experiments, the guidance schedule is fully determined by two parameters: the inference delay $d$ and the ramp length $r$.
Once these two parameters are specified, the corresponding guidance schedule is uniquely defined.
This section details how the delay parameter $d$ is constructed and controlled in our experiments.

All experiments are conducted using the $\pi_{0.5}$ model on a single RTX~4090 GPU.
Without enabling inference-time optimizations, a single forward pass of the model takes approximately 170\,ms.
We adopt an action chunk size of 60, where each chunk corresponds to 2 seconds of continuous actions.
Under this setting, the minimum delay induced by inference latency corresponds to approximately 6 timesteps.

Through careful empirical measurements, we observe that when running on the same hardware, the inference delay remains stable across executions and does not exhibit large fluctuations, typically staying within a one-timestep variation.
To ensure experimental consistency and precise control over the delay parameter, we explicitly construct the effective delay duration.
Specifically, after generating an action chunk, if the actual inference time does not occupy the prescribed number of delay timesteps, we introduce additional idle time to ensure that the total delay equals the target value.

In the main experiments, we fix the delay to $d=8$ timesteps.
This corresponds to an effective delay of approximately 266.7\,ms.
For the delay ablation study, we evaluate three different delay settings with $d \in \{6, 8, 10\}$ timesteps, corresponding to delays of approximately 200\,ms, 266.7\,ms, and 333.3\,ms, respectively.

We emphasize that this explicit construction of delay is introduced solely to control experimental variables and ensure fair comparison across different settings.
Our experiments show that the proposed method maintains strong performance across a range of delay values.
In practical real-world deployments, the delay does not need to be fixed and can instead be handled using a delay buffer, similar to the strategy adopted in Real-Time Chunking (RTC), allowing the guidance schedule to adapt dynamically to runtime conditions.

\subsection{\textbf{Experiments Details}}

We clarify the experimental protocol for the \emph{pour} task.
The main experiments and the ablation studies were conducted with a time gap of approximately one month.
To ensure that potential changes in the environment or updates to the robot system did not affect the reported results, experiments with identical settings to the main experiments were re-run during the ablation phase to enable fully fair comparisons.

Specifically, the main experiments reported in table~1 and table~2, together with the preliminary study shown in \cref{tab:oneshot}, were conducted in the same experimental batch.
The remaining ablation experiments were performed at a later time.
By re-evaluating overlapping settings, we ensure that all reported comparisons reflect methodological differences rather than changes in the experimental setup.

\begin{table}[t]
\centering
\caption{Task completion scoring schemes for all five tasks.
Positive scores are awarded for completing task-relevant steps, while penalties are applied for execution errors.
Each penalty item is capped at a maximum deduction of 3 points per trial.}
\label{tab:task_scoring}
\small
\renewcommand{\arraystretch}{1.15}
\setlength{\tabcolsep}{8pt}

\begin{tabular}{l l r}
\toprule
\textbf{Task} & \textbf{Scoring Item} & \textbf{Score} \\
\midrule
\multirow{4}{*}{Bowl} 
 & Successfully stack one bowl & {\textbf{+2}} \\
 & Bowl tipping or falling & {$-1$} \\
 & Empty grasp & {$-1$} \\
 & Grasping an already stacked bowl & {$-1$} \\
\midrule
\multirow{4}{*}{Pour} 
 & Complete one pouring operation & {\textbf{+$\mathbf{\frac{10}{3}}$}} \\
 & Bowl tipping or falling & {$-1$} \\
 & Empty grasp & {$-1$} \\
 & Blocks spilled outside the bowl & {$-1$} \\
\midrule
\multirow{4}{*}{PickPlace} 
 & All three objects placed into the box & {\textbf{+10}} \\
 & Object dropped & {$-1$} \\
 & Empty grasp & {$-1$} \\
 & Object not placed into the container & {$-1$} \\
\midrule
\multirow{4}{*}{Drawer} 
 & Successfully open the drawer & {\textbf{+10}} \\
 & Pushing the drawer cabinet & {$-1$} \\
 & Empty grasp & {$-1$} \\
 & Incorrect pulling direction & {$-1$} \\
\midrule
\multirow{2}{*}{Towel} 
 & Complete the first fold & {\textbf{+5}} \\
 & Complete the second fold & {\textbf{+5}} \\
\bottomrule
\end{tabular}
\end{table}

\subsection{\textbf{Metric Details}}

\subsubsection{\textbf{Task Completion Score}}

Trajectory smoothness is only one of several factors that influence a model’s final task performance.
Whether a task can be successfully completed also depends on factors such as the generalization of the training data, the consistency between the deployment environment and the data collection setup, and the overall quality of model training.
As a result, using a single binary success rate is insufficient to fully characterize model performance, especially for long-horizon manipulation tasks.

In long-horizon settings, early execution errors can propagate and significantly affect subsequent actions.
Under such conditions, a binary success metric fails to reflect partial progress or distinguish between qualitatively different failure modes.
To more accurately measure task performance, we introduce a task completion score that provides graded feedback based on the extent to which task objectives are achieved.

For each task, we define a structured scoring scheme in which completing meaningful intermediate steps yields positive scores, while execution errors incur penalties.
The scoring design follows two principles.
First, executions that complete more task-relevant steps receive higher scores than those completing fewer steps.
Second, trajectories that complete the task with recoverable errors receive higher scores than those that fail to complete the task, but lower scores than trajectories that complete the task without errors.

We design task-specific completion criteria and penalty rules for all five tasks to ensure that the resulting scores consistently reflect execution quality and task progress, rather than relying solely on a binary notion of success or failure, as shown in \cref{tab:task_scoring}.

\subsubsection{\textbf{Smoothness Metrics}}

We evaluate trajectory smoothness using three complementary metrics that capture different aspects of execution quality: NSPARC, NLDLJ, and overlap RMSE.
All metrics are reported such that \emph{smaller values indicate smoother trajectories}.
For clarity, NSPARC and NLDLJ are defined as the negations of SPARC and LDLJ, respectively.

\paragraph{\textbf{NSPARC (Negative SPARC)}}
SPARC (Spectral Arc Length) measures smoothness in the frequency domain by quantifying the arc length of the normalized velocity magnitude spectrum.
Given a scalar velocity signal $v(t)$ sampled at interval $\Delta t$, we first compute its discrete Fourier transform and obtain the magnitude spectrum $|V(\omega)|$.
The spectrum is normalized by its DC component,
\begin{equation}
\hat{V}(\omega) = \frac{|V(\omega)|}{|V(0)|}.
\end{equation}

An adaptive cutoff frequency $\omega_c$ is selected as the smallest frequency at which $\hat{V}(\omega)$ falls below a predefined threshold, bounded by a maximum cutoff.
The frequency axis is normalized as
\begin{equation}
\tilde{\omega} = \frac{\omega}{\omega_c},
\end{equation}
and the spectral arc length is computed as
\begin{equation}
\mathrm{SPARC}(v) =
- \int_{0}^{\omega_c} 
\sqrt{
\left( \frac{d \tilde{\omega}}{d\omega} \right)^2
+
\left( \frac{d \hat{V}(\omega)}{d\omega} \right)^2
}
\, d\omega.
\end{equation}

We report the negated quantity
\begin{equation}
\mathrm{NSPARC} \triangleq - \mathrm{SPARC},
\end{equation}
such that smaller NSPARC values correspond to smoother trajectories.

For multi-dimensional end-effector trajectories, SPARC is computed separately for translational and rotational motion.
Translational NSPARC is computed using the Euclidean norm of the 3D linear velocity, while rotational NSPARC is computed using the magnitude of the angular velocity after unwrapping the rotation representation.
The final NSPARC score is obtained by averaging over all end-effectors and motion types.

NSPARC primarily captures the distribution of motion energy across frequencies.
Trajectories with oscillations, hesitation, or frequent corrective motions introduce higher-frequency components and yield larger NSPARC values, whereas smooth, continuous motions concentrate energy in low frequencies and result in smaller NSPARC values.

\paragraph{\textbf{NLDLJ (Negative LDLJ)}}
LDLJ (Log Dimensionless Jerk) is a time-domain smoothness metric that penalizes rapid changes in acceleration.
Given a trajectory of duration $T$ with scalar velocity $v(t)$ and scalar jerk $j(t)$, LDLJ is defined as
\begin{equation}
\mathrm{LDLJ} =
- \log \left(
\frac{T^5}{v_{\mathrm{peak}}^2}
\int_{0}^{T} \| j(t) \|^2 \, dt
\right),
\end{equation}
where $v_{\mathrm{peak}} = \max_t |v(t)|$ is the peak velocity.

We report the negated quantity
\begin{equation}
\mathrm{NLDLJ} \triangleq - \mathrm{LDLJ},
\end{equation}
so that smaller NLDLJ values indicate smoother motion.

For multi-dimensional trajectories, jerk is computed by successively differentiating position or rotation vectors to obtain vector jerk, followed by taking the Euclidean norm.
NLDLJ is computed separately for translational and rotational motion, and the final score is averaged across all end-effectors.
To avoid artificially inflated jerk values at chunk boundaries, jerk samples corresponding to chunk connection points are excluded from the computation.

NLDLJ measures smoothness in terms of higher-order temporal continuity.
Trajectories with abrupt acceleration changes or sharp corrective motions yield larger NLDLJ values, while trajectories with gradual acceleration profiles achieve smaller NLDLJ values.

\paragraph{\textbf{Overlap RMSE}}
Overlap RMSE directly measures consistency across consecutive action chunks.
Let $\mathbf{a}^{(k)}_{1:H}$ and $\mathbf{a}^{(k+1)}_{1:H}$ denote two consecutive predicted action chunks of length $H$, and let the last $O$ steps of $\mathbf{a}^{(k)}$ overlap with the first $O$ steps of $\mathbf{a}^{(k+1)}$.
The overlap RMSE is defined as
\begin{equation}
\mathrm{RMSE}_{\mathrm{overlap}} =
\sqrt{
\frac{1}{O}
\sum_{i=1}^{O}
\left\|
\mathbf{a}^{(k)}_{H-O+i}
-
\mathbf{a}^{(k+1)}_{i}
\right\|_2^2
}.
\end{equation}

Overlap RMSE explicitly measures inter-chunk consistency.
Lower overlap RMSE values indicate better alignment between consecutive chunks and smoother continuation behavior at chunk boundaries.

\begin{figure}[t]
\centering
\includegraphics[width=1.0\linewidth]{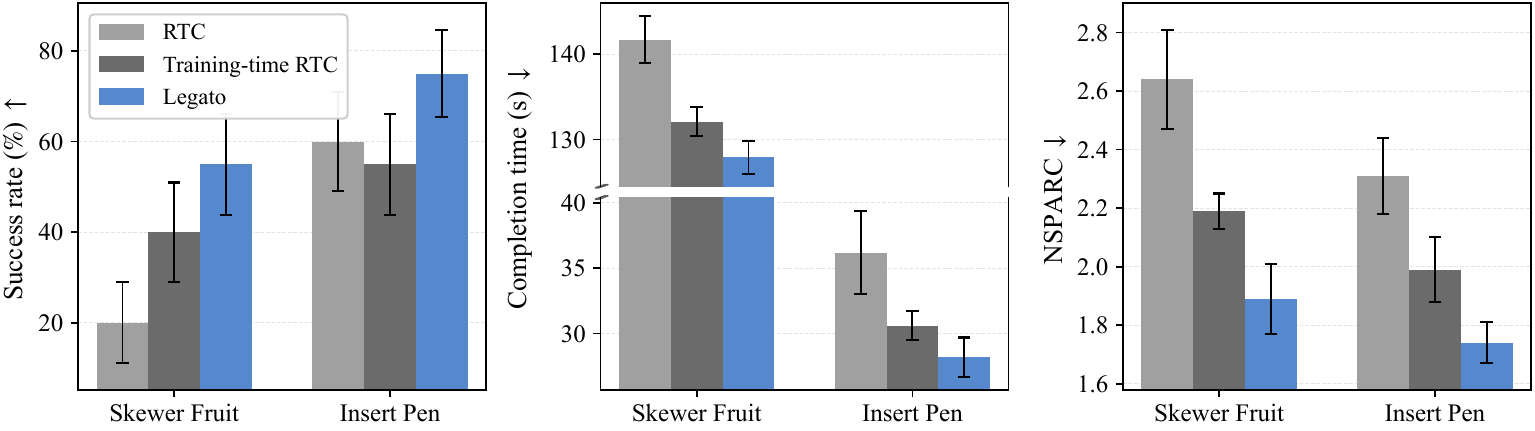}
\vspace{-16pt}
\caption{Legato outperforms RTC and Training-time RTC on two new precision tasks.}
\vspace{-6pt}
\end{figure}


\begin{table}[t]
\centering
\caption{Ablation results comparing one-shot guidance and Legato under the same guidance configuration ($d{=}8$, $s{=}30$, $r{=}22$).
Values are reported as mean $\pm$ standard error.}
\label{tab:oneshot}

\small
\renewcommand{\arraystretch}{0.9}
\setlength{\tabcolsep}{10pt}

\begin{tabular}{l|cc}
\toprule
\textbf{Metric}
& One-shot Guidance
& \cellcolor{legatoblue}{Legato} \\
\midrule

Completion Time $\downarrow$
& \pmstd{88.44}{1.67}
& \cellcolor{legatoblue}\textbf{\pmstd{75.73}{1.51}} \\

NSPARC $\downarrow$
& \pmstd{1.77}{0.17}
& \cellcolor{legatoblue}\textbf{\pmstd{1.65}{0.08}} \\

NLDLJ $\downarrow$
& \pmstd{40.69}{0.21}
& \cellcolor{legatoblue}\textbf{\pmstd{39.50}{0.13}} \\

Overlap RMSE $\downarrow$
& \pmstd{12.69}{1.55}
& \cellcolor{legatoblue}\textbf{\pmstd{5.14}{0.17}} \\

\bottomrule

\end{tabular}

\vspace{-2mm}
\end{table}

\subsection{\textbf{Results on More Challenging Tasks}}

We selected two other tasks: Skewer Fruit and Insert Pen.
Both require fine manipulation and smoothness, as jitter easily causes failure.
Legato shows advantages in time and smoothness, with improved success rates (20 trials per method).
Success rate is not the sole metric: improvements in smoothness and mode-switching suppression benefit low-level control stability and execution time, both critical for execution quality.

\subsection{\textbf{Experiment on the Smoothness–Reactivity Trade-off}}

\noindent\begin{minipage}[t]{0.56\linewidth}
\vspace{0pt}
{We design a reactive task: when the gripper is about to reach the ball, we pull the ball via a rope along the direction perpendicular to the gripper's motion with fixed speed.
We select $d{=}8$, $s{=}8$, $r{=}22$ to achieve higher inference frequency, enabling fair comparison of reactivity and smoothness between Legato and RTC.
Results show similar reactivity but significant smoothness improvement, demonstrating that Legato's smoothness gains do not compromise reactivity.
}
\end{minipage}
\hfill
\begin{minipage}[t]{0.42\linewidth}
\vspace{0pt}
\centering

\includegraphics[width=0.85\linewidth]{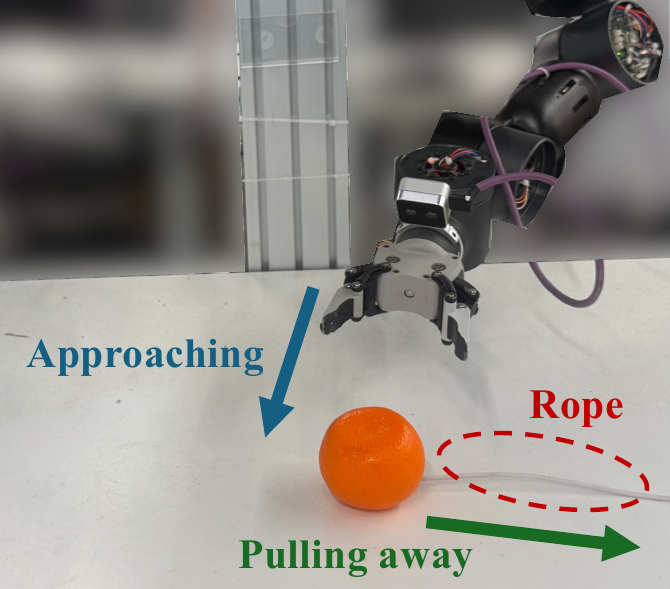}
\captionof{figure}{Reactive task setup.}

\vspace{4pt}
\captionof{table}{Reactivity trade-off.}
{\scriptsize
\setlength{\tabcolsep}{2pt}
\begin{tabular}{lcc}
\toprule
Method & Succ.$\uparrow$ & NSPARC$\downarrow$\\
\midrule
RTC    & 0.7 & $1.97{\pm}0.14$ \\
Legato & 0.7 & $1.48{\pm}0.18$ \\
\bottomrule
\end{tabular}
}

\end{minipage}

\subsection{\textbf{Preliminary Study Details}}
\label{sec:appendix_prelim}

This appendix provides additional empirical evidence supporting the conclusion in section. III-B that one-shot guidance is insufficient for effective continuation and that guidance must be applied before every denoising step.

We compare a one-shot guidance baseline with Legato on the \emph{pour} task.
Both methods are evaluated under the same experimental conditions as the main experiments.
For the one-shot baseline, guidance is applied only at initialization, after which standard multi-step denoising is performed without any intermediate guidance.
Legato, in contrast, applies guidance before every denoising step while remaining consistent with the training objective.

We use a guidance schedule with stride $s=30$, delay $d=8$, and ramp length $r=22$ for both methods.
Quantitative results are reported in \cref{tab:oneshot}.
The results show that Legato significantly outperforms the one-shot baseline, with particularly pronounced improvements in overlap RMSE.
This indicates that without repeated guidance, the overlap region progressively deviates from the desired continuation, even when the initial condition is properly constrained.

These results empirically confirm the observation in section. III-B that guidance applied only at initialization cannot reliably preserve constraints throughout the denoising process.
Repeated, per-step guidance is necessary to maintain consistent continuation across action chunks.

\begin{table}[t]
\centering
\caption{Hyperparameter configuration used in the main experiments and ablation studies.}
\label{tab:hyperparams}
\small
\renewcommand{\arraystretch}{1.2}
\setlength{\tabcolsep}{10pt}

\begin{tabular}{l l l}
\toprule
\textbf{Symbol} & \textbf{Description} & \textbf{Value} \\
\midrule
$H$ & Action chunk size & $60$ \\
$f$ & Action execution frequency & $30$\,Hz \\
$N$ & Number of denoising steps & $5$ \\
$d \sim \mathrm{Uni}[\cdot]$ & Training-time delay range & $\mathrm{Uni}[0,\,10]$ \\
$r \sim \mathrm{Uni}[\cdot]$ & Training-time ramp range & $\mathrm{Uni}[0,\,50]$ \\
\bottomrule
\end{tabular}
\end{table}

\subsection{\textbf{Robot Hardware Configuration}}

All experiments are conducted on the same dual-arm robotic platform.
The robot is equipped with a left arm and a right arm, where each arm consists of seven actuated joints and a gripper, resulting in eight degrees of freedom per arm.

The perception system includes one head-mounted RGB camera providing a global view of the workspace, as well as one wrist-mounted RGB camera on each arm.
In total, the robot uses three cameras for visual observation.

For VLA training and inference, actions are represented in the end-effector space.
Each arm’s action consists of a 6-dimensional end-effector pose, including 3D position and 3D rotation vector, together with a 1-dimensional gripper command.
As a result, the full action space for the dual-arm system is 14-dimensional.

\subsection{\textbf{Hyperparameter Configuration}}

Table~\ref{tab:hyperparams} summarizes the hyperparameter configuration used in the main experiments and ablation studies.
Unless otherwise specified, all experiments share the same configuration.
We note that $d$ and $r$ denote the delay and ramp-length parameters of the guidance schedule; they are fixed at evaluation time, while during training we optionally randomize them by uniform sampling within specified ranges.

\begin{figure}[t]
    \centering
    \includegraphics[width=0.9\linewidth]{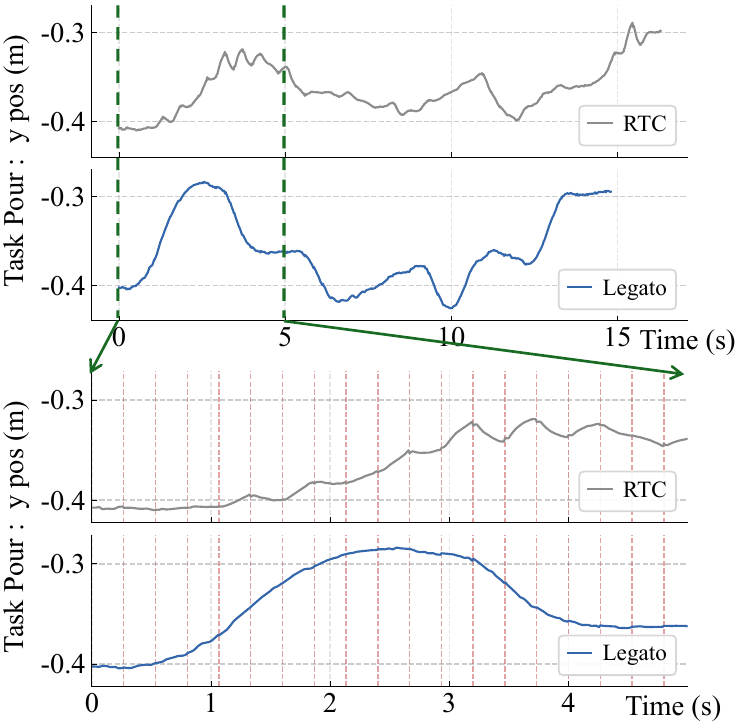}
    \caption{
Trajectory example for the \emph{pour} task under the $d{=}s{=}r{=}8$ configuration.
The top panels show a full pouring operation, and the bottom panels show a zoomed-in view of the first 5 seconds.
Red dashed lines indicate chunk boundaries.
RTC exhibits pronounced low-frequency, large-amplitude oscillations, with direction changes occurring mostly within chunks, indicating increased spurious multimodal switching.
Despite achieving lower overlap RMSE, RTC produces visibly less smooth trajectories in this regime.
}
    \label{fig:abnormal}
\end{figure}

\subsection{\textbf{Results Analysis}}


\subsubsection{\textbf{Analysis of the d=s=r=8 setting}}

We analyze an abnormal behavior observed under the $d{=}s{=}r{=}8$ configuration, where RTC exhibits counterintuitive trends on specific smoothness metrics.

Under this setting, the model initiates the next inference immediately after generating each action chunk.
When the delay $d$ is fixed, setting $s{=}r{=}d$ corresponds to the highest possible inference frequency.

In this regime, the oscillatory behavior of RTC becomes visually apparent, with motion fluctuations reaching amplitudes that are clearly observable.
To better understand this phenomenon, we visualize the executed trajectories in \cref{fig:abnormal}.
The plots reveal large-amplitude oscillations in RTC trajectories, whereas Legato produces substantially smoother motion.

The lower panels show a zoomed-in view of the first 5 seconds of execution.
The vertical red dashed lines indicate chunk boundaries.
Notably, most direction changes occur \emph{within} individual chunks rather than at chunk boundaries.
This suggests that under frequent re-inference, RTC suffers from more severe spurious multimodal switching inside each chunk, rather than discontinuities caused purely by chunk transitions.

Under this setting, NSPARC more faithfully reflects the perceived smoothness difference between RTC and Legato.
Since NSPARC captures the spectral distribution of motion energy, it is particularly sensitive to low-frequency, large-amplitude oscillations, which dominate RTC trajectories in this regime.
In contrast, Legato suppresses such oscillatory behavior by maintaining stronger mode persistence across denoising steps, as shown in \cref{fig:abnormal} and fig. 6.

Interestingly, RTC achieves a lower overlap RMSE than Legato in this configuration.
This observation indicates that overlap RMSE may fail to fully capture smoothness degradation when oscillations are dominated by low-frequency, large-amplitude motion.
Although the overlap between consecutive chunks remains numerically consistent, the resulting trajectory still exhibits pronounced oscillations that negatively impact execution quality.
This case highlights a limitation of overlap RMSE as a standalone smoothness indicator under high-frequency inference settings.

\subsubsection{\textbf{Analysis of the condition row}}

We further analyze the effect of introducing the condition row in the guidance schedule.
As shown in table~IV, adding the condition row does not lead to a significant improvement in NSPARC, whereas it consistently yields a substantial reduction in overlap RMSE across different guidance configurations.
This suggests that the condition row primarily improves inter-chunk consistency rather than intra-chunk smoothness.

When the delay $d$ decreases and the ramp length $r$ correspondingly increases due to parameter constraints, the overlap RMSE of models \emph{without} the condition row also decreases.

Although adding the condition row provides clear benefits under identical $(d,s,r)$ configurations, we observe that a model without the condition row under $(d,s,r)=(6,30,24)$ achieves better overlap RMSE than a model with the condition row under $(d,s,r)=(10,30,20)$.
This observation suggests that when the delay $d$ is sufficiently small, acceptable continuation behavior can be achieved even without the condition row.
In such regimes, omitting the condition row may serve as a viable alternative with reduced conditioning overhead.

\end{document}